\documentclass{article}

\usepackage[final]{nips_2018}

\usepackage{color}
\usepackage[normalem]{ulem}
\usepackage{url}

\usepackage{amsmath}
\usepackage{nicefrac}
\usepackage{natbib}
\setcitestyle{square, numbers}
\usepackage{graphicx}
\usepackage{subcaption}
\usepackage{algorithm}
\usepackage{balance}
\usepackage[noend]{algpseudocode}
\usepackage[colorinlistoftodos]{todonotes}
\usepackage[T1]{fontenc}
\usepackage{booktabs}       
\usepackage{amsfonts}       
\usepackage{microtype}      
\usepackage{float}
\newfloat{algorithm}{t}{lop}

\usepackage{hyperref}
\usepackage{url}

\usepackage[english]{babel}
\usepackage[utf8]{inputenc}
\usepackage{amssymb}
\usepackage{amsthm}
\usepackage{wrapfig}

\newcommand{\bs}[1]{\boldsymbol{#1}}
\newcommand{\E}{\mathds{E}}

\renewcommand{\a}[0]			{ {\bs{\theta}} }

\renewcommand{\vec}[1]			{ \bs{#1} }

\DeclareUrlCommand\ULurl{%
  \renewcommand\UrlLeft{\uline\bgroup}%
  \renewcommand\UrlRight{\egroup}}

\DeclareMathOperator*{\argmax}{\mathrm{argmax}}
\DeclareMathOperator*{\argmin}{\mathrm{argmin}}

\DeclareMathOperator*{\KL}{\mathrm{KL}}
\newcommand*\diff{\mathop{}\!\mathrm{d}}

\usepackage{amsmath,amsfonts,bm}

\def\eqref#1{Equation~(\ref{#1})}

\def\1{\bm{1}}

\DeclareMathAlphabet{\mathsfit}{\encodingdefault}{\sfdefault}{m}{sl}
\SetMathAlphabet{\mathsfit}{bold}{\encodingdefault}{\sfdefault}{bx}{n}

\renewcommand{\E}{\mathbb{E}}

\usepackage{hyperref}
\usepackage{url}

\title{Relative Entropy Regularized Policy Iteration}

\author{Abbas Abdolmaleki, Jost Tobias Springenberg, Jonas Degrave, Steven Bohez, Yuval Tassa,\\ \textbf{Dan Belov, Nicolas Heess, Martin Riedmiller} \\
DeepMind, London, UK \\
\texttt{\{aabdolmaleki,springenberg,grave,sbohez,danbelov,heess,riedmiller\}@google.com}
}

\begin{document}

\maketitle
\begin{abstract}
We present an off-policy actor-critic algorithm for Reinforcement Learning (RL) that combines ideas from gradient-free optimization via stochastic search with learned action-value function.
The result is a simple procedure consisting of three steps: i) policy evaluation by estimating a parametric action-value function; ii) policy improvement via the estimation of a local non-parametric policy; and iii) generalization by fitting a parametric policy. Each step can be implemented in different ways, giving rise to several algorithm  variants.
Our algorithm draws on connections to existing literature on black-box optimization and ``RL as an inference'' and it can be seen either as an extension of the Maximum a Posteriori Policy Optimisation algorithm (MPO) \citep{abdolmaleki2018maximum}, or as an extension of {\it Trust Region} Covariance Matrix Adaptation Evolutionary Strategy (CMA-ES)  \citep{abdolmaleki2017TRCMA, HansenCMAES} to a policy iteration scheme. Our comparison on 31 continuous control tasks from parkour suite \citep{heess2017emergence}, DeepMind control suite \citep{ControlSuite} and OpenAI Gym \citep{brochmanopenaigym} with diverse properties, limited amount of compute and a single set of hyperparameters, demonstrate the effectiveness of our method and the state of art results. Videos, summarizing results, can be found at \href{https://sites.google.com/corp/view/aistat/home}{\ULurl{goo.gl/HtvJKR}}. 
\end{abstract}

\section{Introduction}
Reinforcement learning with flexible function approximators such as neural networks, also referred to as ``deep RL'', holds great promises for continuous control and robotics. Neural networks can express complex dependencies between high-dimensional and multimodal input and output spaces, and learning-based approaches can find solutions that would be difficult to craft by hand.
Unfortunately, the generality and flexibility of learning based approaches with neural networks can come at a price: Deep reinforcement learning algorithms can require large amounts of training data; they can suffer from stability problems, especially in high-dimensional continuous action spaces~\citep{duan2015benchmarking,schulman15}; and they can be sensitive to hyperparameter settings. Even though attempts to control robots or simulated robots with neural networks go back a long time~\citep{Pomerleau89,RiedmillerDrive,Stone2005}, it has only been recently that algorithms have emerged which are able to scale to challenging problems~\citep{bansal2018emergent,heess2017emergence,openaidex} -- including first successes in the data-restricted domain of physical robots~\citep{Gu17,kalashnikov2018qtopt,openaidex,riedmiller2018learning}.

Model-free off-policy actor-critic algorithms have several appealing properties. In particular, they make minimal assumptions about the control problem, and can be data-efficient when used in combination with a appropriate data reuse. They can also scale well when implemented appropriately (see e.g.~\citet{Gu17,Popov17}). Broadly speaking, many off-policy algorithms are implemented by alternating between two steps: i)~\textbf{a policy evaluation step} in which an action-value function is learned for the current policy; and ii)~\textbf{a policy improvement step} during which the policy is modified given the current action-value function. 


In this paper we outline a general policy iteration framework and motivate it from both an intuitive perspective as well as a “RL as inference” perspective. In the case when the MDP collapses to a bandit setting our framework can be related to the black-box optimization literature. We propose an algorithm that works reliably across a wide range of tasks and requires minimal hyper-parameter tuning to achieve state of the art results on several benchmark suites. Similarly to Maximum a Posteriori Policy Optimisation algorithm (MPO) ~\citep{abdolmaleki2018maximum}, it estimates the action-value function $Q^{\pi}(s,a)$ for a policy $\pi$ and then uses this Q-function to update the policy. The policy improvement step builds on ideas from the black-box optimization and KL-regularized control literature. It first estimates a local, non-parametric policy that is obtained by reweighting the samples from the current/prior policy, and subsequently fits a new parametric policy via weighted maximum likelihood learning. Trust-region like constraints ensure stability of the procedure.
The algorithm simplifies the original formulation of MPO while improving its robustness via decoupled optimization of policy's mean and covariance. We show that our algorithm solves standard continuous control benchmark tasks from the DeepMind control suite (including control of a humanoid with 56 action dimensions), from the OpenAI Gym, and also the challenging ``Parkour'' tasks from \citet{heess2017emergence}, all with the same hyperparameter settings and a single actor for data collection.

\section{Problem Statement}
\label{sec:problem}
In this paper we are focused on actor-critic algorithms for stable and data efficient policy optimization. Actor-critic algorithms decompose the policy optimization problem into two distinct sub-problems as also outlined in Algorithm \ref{alg:MPONON}: i) estimating the state-conditional action value (the Q-function, denoted by $Q$) given a policy $\pi$, and, ii) improving $\pi$ given an estimate of $Q$. 

We consider the usual discounted reinforcement learning (RL) problem defined by a Markov decision process (MDP). The MDP consists of continuous states $s$, actions $a$, an initial state distribution $p(s_0)$, transition probabilities $p(s_{t+1} | s_t, a_t)$ which specify the probability of transitioning from state $s_t$ to $s_{t+1}$ under action $a_{t}$, a reward function $r(s, a) \in \mathbb{R}$ and the discount factor $\gamma \in [0, 1)$. The policy $\pi(a | s, \a)$ with parameters $\a$ is a distribution over actions $a$ given a state $s$. We optimize the objective, 
\begin{equation}
\textstyle{J(\pi) = \E_{\pi,p(s_0)} \Big\lbrack \sum_{t=0}^\infty \gamma^t r(s_t, a_t) | s_0 \sim p(\cdot), a_t \sim \pi(\cdot | s) \Big\rbrack} \label{eq:mainobj}
\end{equation}

where the expectation is taken with respect to the trajectory distribution induced by $\pi$.
We define the action-value function associated with $\pi$ as the expected cumulative discounted return when choosing action $a$ in state $s$ and acting subsequently according to policy $\pi$ as $Q^\pi(s,a) = \E_\pi \lbrack \sum_{t=0}^\infty \gamma^t r(s_t, a_t) | s_0=s, a_0 = a]$. This function satisfies the recursive expression $Q(s_t, a_t) = \E_{p(s_{t+1} | s_t, a_t)} \big[ r(s_t, a_t) + \gamma V(s_{t+1}) \big]$ where $V^\pi(s) = \E_\pi[ Q^\pi(s,a) ]$ is the value function of $\pi$.

The true Q-function of an MDP and policy $\pi^{(k)}$ in iteration $k$ provides the information needed to estimate a new policy $\pi^{(k+1)}$ that will have a higher expected discounted return than $\pi^{(k)}$; and will thus improve our objective. This is the core idea underlying policy iteration~\citep{Sutton1998}: if for all states we change our policy to pick actions that have higher value with higher probability then the overall objective is guaranteed to improve. 
For instance, we could attempt to choose $\pi^{(k+1)}(a | s) = \delta( a^*(s) - a)$ where $a^*(s) = \argmax_{a} Q(s,a)$ as action selection rule -- assuming an accurate $Q^{\pi}$. 

In this paper we specifically focus on the problem of reliably optimizing $\pi^{(k+1)}$ given $Q^{\pi^{(k)}}$. In particular, in Section \ref{sec:algorithm} we discuss update rules for stochastic policies that explicitly control the change in $\pi$ from one iteration to the next. And we show how to avoid premature convergence when Gaussian policies are used.
\begin{algorithm}[t]
\caption{Actor-Critic}\label{alg:MPONON}
\begin{algorithmic}
\renewcommand{\thealgorithm}{}
\State {Initialize $\pi^{(0)} $, $Q^{\pi^{(-1)}}$, $k \leftarrow 0$}
\Repeat 
\State{$Q^{\pi^{(k)}} \leftarrow \mathrm{PolicyEvaluation}(\pi^{(k)}, Q^{\pi^{(k-1)}})$}  
\State{$\pi^{(k+1)} \leftarrow \mathrm{PolicyImprovement}(\pi^{(k)}, Q^{\pi^{(k)}})$} 
\State{$k \leftarrow k+1$}
\Until{convergence}
\end{algorithmic}
\end{algorithm}


\section{Policy Evaluation (Step 1)}
Policy Evaluation is concerned with learning an approximate Q-function (policy evaluation). In principle, any off-policy method for learning Q-functions could be used here, as long as it provides sufficiently accurate value estimates. This includes making use of recent advances such as distributional RL \citep{BellemareDM17,d4pg} or Retrace \citep{munos2016safe}. To separate the effects of Policy Improvement and better value estimation we focus on simple 1-step temporal difference (TD) learning for most of the paper (showing advantages from better policy evaluation approaches in separate experiments). We fit a parametric Q-function $Q_\phi^\pi(s, a)$ with parameters $\phi$ by minimizing the squared (TD) error
\begin{equation*}
\textstyle{
    \min_\phi \left(r_t + \gamma Q_{\phi'}^{\pi^{(k-1)}}\left(s_{t+1}, a_{t+1}\sim\pi^{(k-1)}(a|s_{t+1})\right) - Q_\phi^{\pi^{(k)}}\left(s_t, a_t\right)\right) ^2,
}
\end{equation*}
where $r_t = r(s_t, a_t)$, which we optimize via gradient descent. We let $\phi'$ be the parameters of a target network (the parameters of the last Q-function) that is held constant for $250$ steps (and then copied from the optimized parameters $\phi$). For brevity of notation we drop the subscript and dependence on parameters $\phi$ in the following section and write $Q^{\pi^{(k)}}(a, s)$.

\section{Policy Improvement (Step 2-3)}
\label{sec:algorithm}
The policy improvement step consists of optimizing $\bar{J}(s,\pi) = \E_\pi[ Q^{\pi^{(k)}}(s,a)]$ for $s$ drawn from the visitation distribution $ \mu_\pi(s)$. In practice, we replace $\mu_\pi(s)$ with draws from a replay buffer. As argued intuitively in Section \ref{sec:problem}, if we improve this expectation in all states and for an accurate $Q$, this will improve our objective $J$ (Equation \ref{eq:mainobj}). Below we describe two approaches that perform this optimization. They do not fully optimize $\bar{J}$ to avoid being misled by errors in the approximated Q-function -- while keeping exploration. We find a solution capturing some information about the local value landscape in the shape of the distribution. Maintaining this information is important for exploration and future optimization steps. Both approaches employ a two-step procedure: they first construct a non-parametric estimate $q$ s.t. $\bar{J}(s,q) \geq \bar{J}(s,\pi^{(k)})$ (Step 2). They then project this non-parametric representation back onto the manifold of parameterized policies by finding
\begin{equation}
\textstyle{
  \pi^{(k+1)} = \argmin_{\pi_\a} \mathbb{E}_{\mu_\pi(s)}\Big[\mathrm{KL}\Big(q(a | s) \| \pi_\a(a | s)\Big)\Big] 
}
\label{eq:klmin}
\end{equation}
which amounts to supervised learning -- or maximum likelihood estimation (MLE) (Step 3). This split of the improvement step into sample based estimation followed by supervised learning allows us to separate the neural network fitting from the RL procedure, enabling regularization in the latter.

\subsection{Finding action weights (Step 2)\label{sec:weightcalc}}
Given a learned approximate Q-function, in each policy optimization step, we first sample K states $\{{s_j}\}_{j=1...K}$ from the replay buffer. Secondly, we
sample N actions for each state $s_j$ from the last policy distribution, forming the sample based estimate, i.e,
$
  \{a_{i}\}_{i=1...N} \sim \pi^{(k)}(a | s_j),
$
where i denotes the action index and j denotes the state index in the replay. We then evaluate each state-action pair using the Q-function ($Q^{\pi^{(k)}}$). Now, given states, actions, and their corresponding Q-values, i.e. $\{{s_j,\{a_{i},Q^{\pi^{(k)}}(s_j, a_i)\}_{i=1...N}}\}_{j=1...K}$ we want to first re-adjust the probabilities for the given actions in each state
such that better actions have higher probability. These updated probabilities are expressed via the weights $q_{ij}$, forming the non-parametric, sample based improved policy, i.e,
$
\forall s_j, a_i:  q(a_i | s_j) = q_{ij}.
$
To determine $q_{ij}$, one could assign probabilities manually to the actions based on the ranking of actions w.r.t their Q-values. This approach has been used in the black-box and stochastic search communities and can be related to methods such as CMA-ES~\citep{HansenCMAES} and the cross-entropy method~\citep{Rubinstein2004}. 
In general, \textbf{we can calculate weights using any rank preserving transformation of 
the Q-values}. If the weights additionally form a proper sample based distribution, satisfying: i) positivity of weights, and ii) normalization $\sum_i q_{ij} = 1$. We now discuss various valid transformations of the Q-values.


\paragraph{Using ranking to transform Q-values.}
In particular, one such weighting would be to choose the weight of the $i$-th best action for the $j$-th sampled state to be proportional to $q_{ij} \propto \ln (\frac{N + \eta}{i})$, where N is the number of action samples per state and $\eta$ is a temperature parameter (if $\eta = 0.5$ this would correspond to an update similar to CMA-ES). Intuitively, we set a fixed pseudo probability for each action based on their rank, such that the expected Q-value under this new sample-based (state dependent) distribution increases.

\paragraph{Using an exponential transformation of the Q-values.}
Alternatively, we can obtain the weights by optimizing for an optimal assignment of action probabilities directly. If we additionally want to constrain the change of the policy this corresponds to solving the following KL regularized objective:
\begin{align*}
q_{ij} =&\argmax_{q(a_{i} | s_j)} \sum_j^K \sum_i^N q(a_i | s_j) Q^{\pi^{(k)}}(s_j, a_i) \\ 
 s.t. &\frac{1}{K}\sum_j^K \sum_i^N q(a_i | s_j)\log \frac{q(a_i | s_j)}{\frac{1}{N}} < \epsilon,  \quad \quad \forall_{j} \sum_i^N q(a_i| s_j) = 1.
\end{align*}
Here, the first constraint forces the weights to stay close to the last policy probabilities, i.e. bounds the average relative entropy, or average KL, since samples $a_{i}$ are drawn from $\pi^{(k)}$. The second constraint ensures that weights are normalized. The solution will be new weights, given through the categorical probabilities $q_{ij} = q(a_i | s_j)$, such that the expected Q-value increases while constraining the reduction in entropy (to prevent the weights from collapsing onto one action immediately). This objective has been used in the RL and bandit optimization literature before (see e.g. \citep{Peters10,mpo18}) and, when combined with Q-learning has some optimality guarantees \citep{abdolmaleki2018maximum}. As it turns out, its solution can be obtained in closed form, and consists of a softmax over Q-values:
$$
    q_{ij} = q(a_{i},s_j)  = \exp\left(\nicefrac{Q^{\pi^{(k)}}\left(s_j,a_i\right)}{\eta}\right)/Z(j),
$$
where $Z(j) =\sum_i \exp\Big(\nicefrac{Q^{\pi^{(k)}}(s_j,a_i)}{\eta}\Big)$. The temperature $\eta$ corresponding to the constraint $\epsilon$ can be found automatically by solving the following convex dual function alongside our policy optimization:
\begin{equation*}
\textstyle{
\eta = \argmin_{\eta} \eta\epsilon+\eta\sum_j^K\frac{1}{K}\log\left(\sum_i^N\frac{1}{N}\exp\Big(\frac{Q(s_j,a_i)}{\eta}\Big)\right).
}
\end{equation*}
We found that, in practice, this optimization can be performed via a few steps of gradient descent on $\eta$ for each batch after the weight calculation. As $\eta$ should be positive, we use a projection operator to project back the $\eta$ to feasible positive space after each gradient step. We use Adam \citep{kingma2014adam} to optimize $\eta$ together with all other parameters. We refer to the appendix in section B for a derivation of this objective from RL as Inference perspective and the dual.

\paragraph{Using an identity transformation.}
An interesting other possibility is to use an identity transformation. While not respecting the desiderata from above, this would bring our method close to an expected policy gradient algorithm \citep{EPGCiosek}. We discuss this choice in detail in Section A in the appendix.

\begin{algorithm}[t]
\small
\caption{KL Regularized Policy Improvement}\label{Alg:GradientFree}
\begin{algorithmic}[1]
\State {\bf given} batch-size (N), Number of actions (K), Q-function $Q^{\pi^{(k)}}$(target-network), old-policy $\pi^{(k)}$ (target network) and replay-buffer
\State {\bf initialize $\pi_\a$ from the parameters of $\pi^{(k)}$} 
\Repeat
\State {Sample batch of size N from replay buffer}
\State {\bf // Step 2: sample based policy (weights)}
\State $q(a_i | s_j) = q_{ij}$, {\bf computed as:}
\For{j = 1,...,$K$} 
\For{i = 1,...,$N$}
\State {$a_{i} \sim \pi_{target}(a|s_j)$} 
\State {$Q_{ij} = Q^{\pi^{(k)}}(s_{j}, a_i)$} 
\State $q_{ij} =$ {\bf Compute Weights}($\{Q_{ij}\}_{i=1\dots N}$), see section \ref{sec:weightcalc}
\EndFor
\EndFor
\State {\bf // Step 3: update parametric policy}
\State {Given the data-set $\{s_j,(a_{i},q_{ij})_{i=1...N}\}_{j=1...K}$}
\State {\bf Update the Policy by finding }
\State $\pi^{(k+1)} = \argmax_\theta \sum_j^K \sum_i^N q_{ij} \log \pi_{\a}(a_i|s_j)$
\State {\bf (subject to additional (KL) regularization)}, see section \ref{sec:supervisedstep}
\Until{Fixed number of steps}
\State return $\pi^{(k+1)}$
\end{algorithmic}
\end{algorithm}
\subsection{Fitting an improved policy (Step 3) \label{sec:supervisedstep}}
\label{sect:step2}
So far, for each state, we obtained an improved sample-based distribution over actions. Next, we want to generalize this sample-based solution over state and action space -- which is required when we want to select better actions in unseen situations during control. For this, we solve a weighted supervised learning problem
\begin{equation}
  \pi^{(k+1)} = \argmax_{\pi_\theta} \sum_j^K \sum_i^N q_{ij} \log \pi_{\a}(a_i|s_j),
  \label{eq:mle}
\end{equation}
where $\a$ are the parameters of our function approximator (a neural network) which we initialize from the weights of the previous policy $\pi^{(k)}$.
This objective corresponds to minimization of the KL divergence between the sample based distribution $\hat{\pi}$ from Step 2 and the parametric policy $\pi_\theta$, as given in \eqref{eq:klmin}.

Unfortunately, sample based maximum likelihood estimation can suffer from overfitting to the samples from Step 2. Additionally, these sample weights themselves can be unreliable 
due to a poor approximation of $Q^{\pi^{(k)}}$ -- potentially resulting in a large change of the action distribution in the wrong direction when optimizing \eqref{eq:mle}. One effective regularization that addresses both concerns is to limit the overall change in the parametric policy. This additional regularization has a different effect than enforcing tighter constraints in Step 2, which would still only limit the change in the sample-based distribution. To direcly limit the change in the parametric policy  (even in regions of the action space we have not sampled from) 
we thus employ an additional KL constraint\footnote{We note that other commonly used regularization techniques might be worth investigating.} and change the objective from \eqref{eq:mle} to
\begin{equation}
\begin{aligned}
\pi^{(k+1)} = \argmax_{\pi_\theta} \sum^K_j \sum^N_i q_{ij} \log \pi_{\a}(a_i|s_j), \quad \textrm{s.t.}\:\sum^K_j \frac{1}{K} \KL(\pi^{(k)}(a|s_j) \,\|\, \pi_\a(a|s_j) ) < \epsilon_{\pi},
\end{aligned}
\label{eq:mpo}
\end{equation}
where $\epsilon_\pi$ denotes the allowed expected change over state distribution in KL divergence for the policy.
To make this objective amenable to gradient based optimization we employ Lagrangian Relaxation, yielding the following primal optimization problem:
\begin{align*}
\max_\a \min_{\alpha > 0} L(\a,\eta) = \sum_j\sum_i q_{ij} \log \pi_{\a}(a_{i}|s_j) + \alpha\Big(\epsilon_{\pi} - \sum^K_j \frac{1}{K} \KL(\pi^{(k)}(a|s_j) \,\|\, \pi_\a(a|s_j) )\Big).
\end{align*}
We solve for $\a$ by iterating the inner and outer optimization programs independently: We fix the parameters $\a$ to their current value and optimize for the Lagrangian multipliers (inner minimization) and then we fix the Lagrangian multipliers to their current value and optimize for $\a$ (outer maximization). In practice we found it effective to simply perform one gradient step each in inner and outer optimization for each sampled batch of data. This lead to good satisfaction of the constraints throughout coordinate gradient decent training.

\subsubsection{Fitting an improved Gaussian policy}
\label{sec:gaussian}
\begin{wrapfigure}{R}{0.5\textwidth}
  \centering    
  \includegraphics[width=0.5\textwidth]{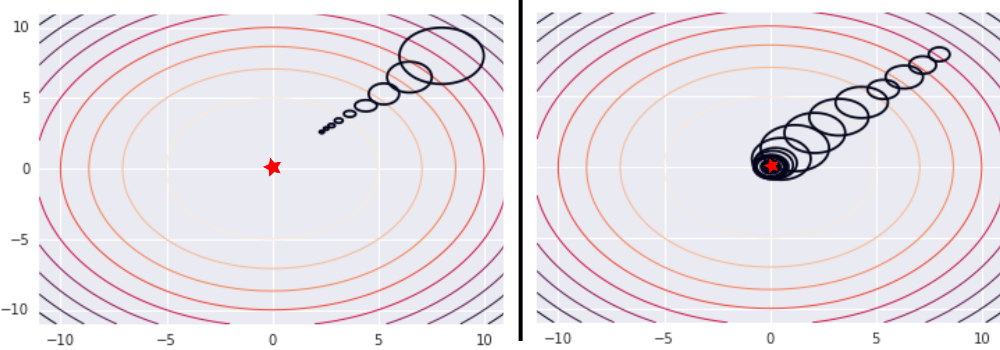}
  \caption{We visualize the optimization of a quadratic Q-function for state [0,0] with maximum likelihood estimation (Equation \ref{eq:mpo}) on the left, and the decoupled updates we propose on the right. MLE leads to premature convergence even with KL constraint on change of the policy. An expanded plot is given in the appendix in Figure 8.}
  \label{fig:spheremain}
  \vspace{-0.3cm}
\end{wrapfigure}
The method described in the main part of Section \ref{sect:step2} works for any distribution. However, in particular for continuous action spaces it still can suffer from premature convergence as it is shown in Figure \ref{fig:spheremain}(left). The reason is that, in each policy improvement step we are essentially optimising for the expected reward for state given actions from the last policy. In such a setting, the optimal solution is to give a probability of 1 to the best action (or equal probabilities to equally good actions) based on its Q-value and zero to other actions. This means that the policy will collapse on the best action to optimise the expected reward even though the best action is not the true optimal action. We can postpone this effect by adding a KL constraint, however, in each iteration the policy will lose entropy to cover the best actions it has seen, albeit slowly, depending on the shape of $Q$ and the choice of $\epsilon_\pi$. And it therefore still can converge prematurely. 

We found that when using Gaussian policies, a simple change can avoid premature convergence in Step~3: we can decouple the objective for the policy mean and covariance matrix which, as intuitively described below, will fix this issue. This technique is also employed in the CMA-ES and TR-CMA-ES algorithms~\cite{HansenCMAES,abdolmaleki2017TRCMA} for bandit problems, but we generalize it to non-linear parameterizations. Concretely, we jointly optimize the neural network weights $\theta$ to maximize two objectives: one for updating the mean with the the covariance fixed to the one of the last policy (target network) and one for updating the covariance while fixing the mean to the one from the target network. This yields the following optimization objectives for the updated mean and covariance:
\begin{align*}
&\pi^{(k+1)} = \argmax_{\mu_\theta,\Sigma_\theta} \sum_j^K \sum_i^N q_{ij} \log \pi_{\a}(a_i|s_j;\Sigma=\Sigma^k) 
+ \sum_j^K \sum_i^N q_{ij} \log \pi_{\a}(a_i | s_j;\mu=\mu^k) \\
\textrm{s.t.}&\: \epsilon_{\mu} > \frac{1}{K}\sum_j^K \KL(\pi^{(k)}(a|s_j) \,\|\, \pi_\a(a|s_j;\Sigma=\Sigma^k) ),\: \\ 
&\epsilon_{\Sigma} > \frac{1}{K}\sum_j^K \KL(\pi^{(k)}(a | s_j) \,\|\, \pi_\a(a | s_j;\mu=\mu^k) ).
\end{align*}
Here, $\mu^k$ and $\Sigma^k$ respectively refer to the mean and covariance of obtained from the previous policy $\pi^{(k)}$ and $\pi_\a(a | s) = \mathcal{N}(\mu_\theta,\Sigma_\theta)$. We solve this optimization by performing gradient descent on an objective derived via the same Langrangian relaxation technique as in Section~\ref{sect:step2}.

\begin{figure*}[t]
\centering
\begin{minipage}[c]{1\textwidth}
\def\mywidth{0.245}
\includegraphics[width=\mywidth\textwidth]{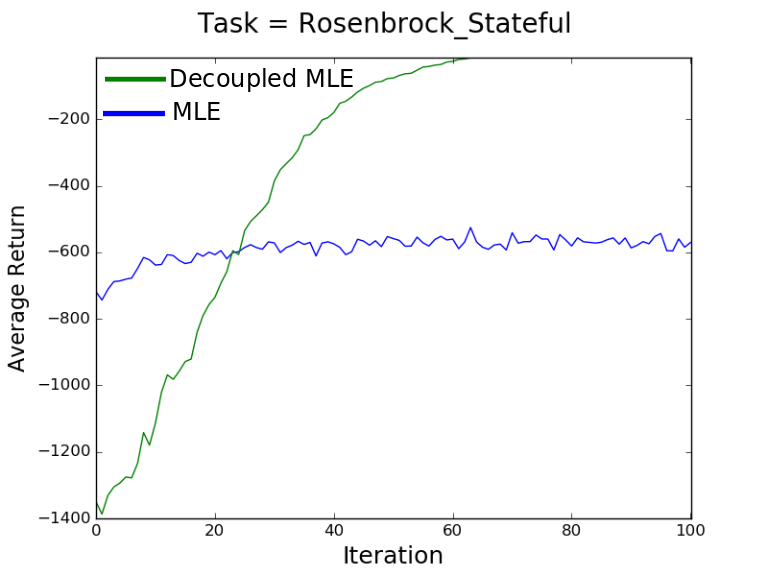}
\includegraphics[width=\mywidth\textwidth]{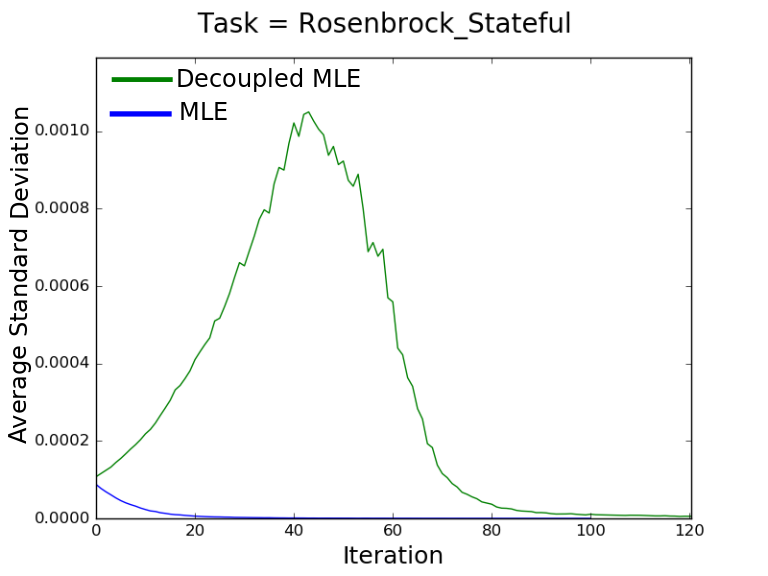}
\includegraphics[width=\mywidth\textwidth]{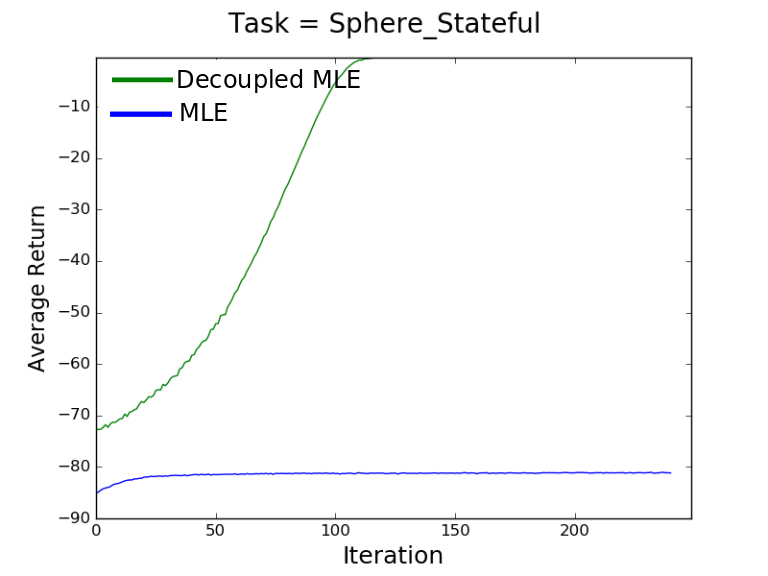}
\includegraphics[width=\mywidth\textwidth]{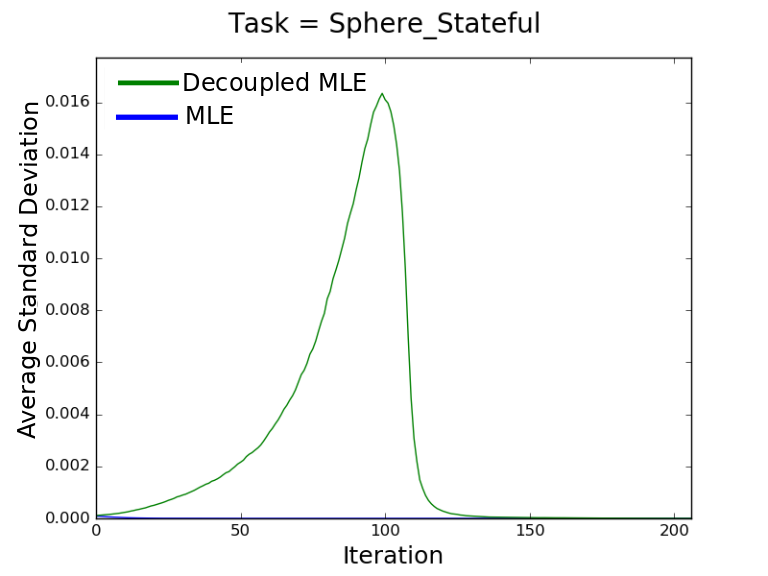}
\caption{Comparison on a statefull Rosenbrock and sphere function with a 10 dimensional state-action space. The results show that the decoupled updates at the beginning naturally increase the variance of the distribution and, after finding the optimum, lead to a decrease in the variance. }
\label{fig:highdim}
\end{minipage}
\vspace{-0.5cm}
\end{figure*}

This procedure has two advantages: 1) the gradient w.r.t. the parameters of the covariance is now independent of changes in the mean; hence the only way the policy can increase the likelihood of good samples far away from the mean is by stretching along the value landscape. This gives us the ability to grow and shrink the distribution supervised by samples without introducing any extra entropy term to the objective \citep{abdolmaleki2015model,tangkaratt2017guide} (see also Figures \ref{fig:spheremain} and \ref{fig:highdim} for an experiment showing this effect). 2) we can set the KL bound for mean and co-variance separately. The latter is especially useful in high-dimensional action spaces, where we want to avoid problems with ill-conditioning of the covariance matrix but want fast learning, enabled by large changes to the mean. The complete algorithm is listed in Algorithm \ref{Alg:GradientFree}.

Please note that the objective we optimise here still is the weighted maximum likelihood objective from Equation \ref{eq:mpo}, with the difference that we optimise it in a coordinate ascent fashion - resulting in the decoupled updates with different KL bounds. In general, such a procedure can also be applied for optimising different policy classes. For other distributions, such as mixtures of Gaussians, we can still use the same procedure and optimise for means, covariances and categorical distribution independently, getting the same effect as for the Gaussian case. If the policy is deterministic (as in DDPG) then the exploration variance is fixed and we would simply optimize the mean of a Gaussian. For categorical distributions each component can be optimized independently. However, the application of the coordinate ascent updates will have to be derived on a per distribution basis. 


\section{Related Work}
\label{sec:ES}
Our algorithm employs ideas used in the family of Evolutionary Strategies (ES) algorithms. The objectives for the Gaussian case in Section \ref{sec:gaussian} can be seen as a generalization of the {\it trust region} CMA-ES updates \citep{HansenCMAES, abdolmaleki2017TRCMA} and similar algorithms 
\citep{wierstra2008fitness,Rubinstein2004}
to a, stateful, sequential setting with an imperfectly estimated evaluation function.
This is discussed further in Section C in the appendix. However, rather than optimizing mean and covariance directly, we assume that 
these are parameterized as a non-linear function of the state, and we combine gradient based optimization of network parameters with gradient-free updates to the policy distribution in action space. Separately previous work has used ES to directly optimize the weights of a neural network policy \citep{ESOpenai} which can be sample inefficient in high-dimensional parameter spaces.
In contrast, our approach operates in action space and exploits the sequentiality of the RL problem.

An alternative view of our algorithm is obtained from the perspective of RL as inference. This perspective has recently received much attention in the literature \citep{GPSLevine,chebotar2016path,PCL17, abdolmaleki2018maximum}
and a number of expectation-maximization based RL algorithms have been proposed (see e.g.  \citep{neumann2011variational,Peters10,Deisenroth2013}) including the original MPO algorithm \citep{abdolmaleki2018maximum}. Concretely, the objectives for policy improvement algorithms mentioned above can be obtained from the perspective of performing Expectation Maximization (EM) on the likelihood $\log p(R | s, a, \a)$ where $R$ denotes an optimality event whose density is proportional to the Q-value. More details on this connection are given in Section B in the appendix. In particular, we can recover MPO by choosing an exponential transformation in the weighting step and removing decoupled updates on mean and covariance. MPO, in turn, is related to relative entropy policy search (REPS) \citep{Peters10}, with differences due to the construction of the sample based policy (REPS considers a sample based approximation to the  joint state-action distribution $p(a, s)$) and the additional regularization in the policy fitting step which is not presented in REPS. 

Conservative policy search algorithms such as Trust Region Policy Optimization~\citep{schulman15}, Proximal Policy Optimization~\citep{schulman2017ppo} and their many derivatives 
make use of a similar KL constraint as in our Step 3 to stabilize learning. The supervised nature of our policy fitting procedure resembles methods from Approximate Dynamic Programming such as Regularized~\citep{Rarahmand09} and  Classification based Policy Iteration \citep{CPI16} -- which has been scaled to high-dimensional discrete problems \citep{ADPTetris13} -- and the classic Cross-Entropy Method (CEM) \citep{Rubinstein2004}. The idea of separating fitting and improvement of the policy is shared with works such as \cite{GPSLevine} and \cite{chebotar2016path,GPS16}.



\section{Results}

To illustrate core features of our algorithm we first present results on two standard optimization problems. These highlight the benefit of decoupled maximum likelihood for Gaussian policies. We then perform experiments on 24 tasks from the DeepMind control suite \citep{ControlSuite}, three high dimensional parkour tasks from \citep{heess2017emergence} and four high dimensional tasks from OpenAI gym \citep{brochmanopenaigym}. Depictions of task sets are in the appendix (Figure 9, 10).

\subsection{Standard Functions}
To isolate the evaluation of our policy improvement procedure from errors in the estimation of $Q^\pi$ we performed experiments using two fixed standard functions.
We make both functions state and action dependent by first defining an auxiliary variable $y = a + s$, that varies linearly with to the action (the functions can thus seen as ``ground truth'' Q-functions. We consider: i) the sphere function $Q(a,s) = -\sum_{i=1}^N {y_i}^2$, and ii) the well known Rosenbrock function~\citep{molga2005test} $Q(a,s) = -\sum_{i=1}^{n-1} [100(y_{i+1} - y_{i}^2)^2 + (1-y_{i}^2)]$. The global optimal action for state $s$ for both of these functions is given as $a^* = -s$; in which the optimal Q-value of zero is obtained. Instead of using a replay buffer, we sample 100 states from a uniform state distribution in the interval $[-2; 2]$ for each batch and sample 10 actions from our current policy for calculating weights. 

The results for the dimensional sphere function are depicted in Figure~\ref{fig:spheremain}. We plot the learning progress of the Gaussian policy for state $[0;0]$ (every 20 iterations) for both the weighted MLE which also used by MPO \cite{abdolmaleki2018maximum} and the decoupled optimization approach. The decoupled optimization starts by increasing the variance. Only when the optimum is found the variance start shrinking, and the distribution successfully converges on the optimum. The MLE procedure always shrinks variance, causing premature convergence
 even though we purposefully started with a larger variance for the MLE objective.

Figure~\ref{fig:highdim} shows the average return over states for each iteration as well as the policy standard deviation for 10 dimensional versions of the Rosenbrock and Sphere functions. We observe that the decoupled optimization successfully solves both tasks, although the initial standard deviation is small. In contrast, the MLE approach converges prematurely. 

\subsection{Continuous control benchmark tasks}
\begin{figure*}[ht]
\centering
\def\mywidth{0.24}
\begin{subfigure}{\mywidth\textwidth}
  \centering
  \includegraphics[width=\textwidth]{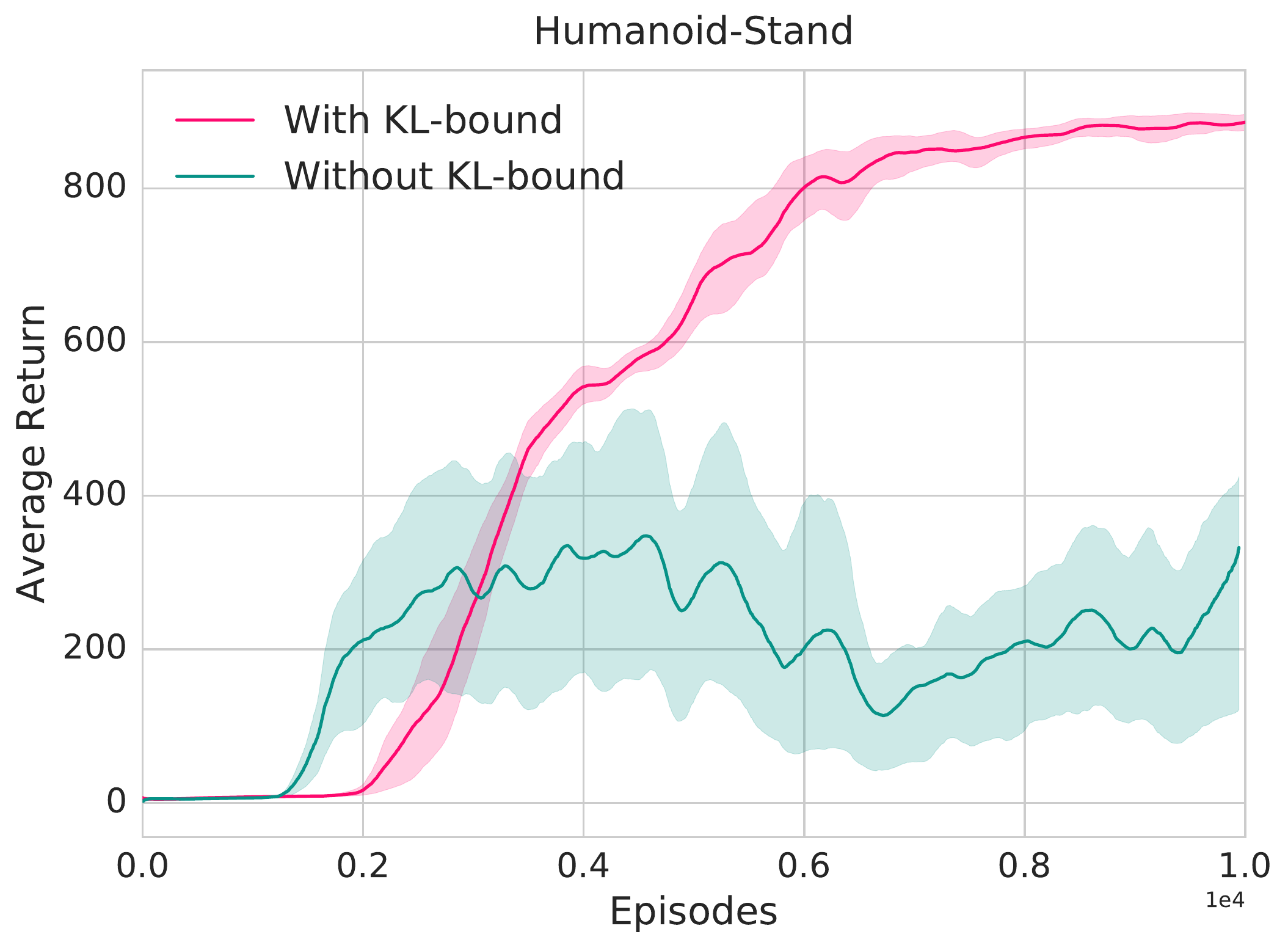}
  \caption{No KL bound }
  \label{fig:meanCovBound:kl}
\end{subfigure}%
\begin{subfigure}{\mywidth\textwidth}
  \centering
  \includegraphics[width=\textwidth]{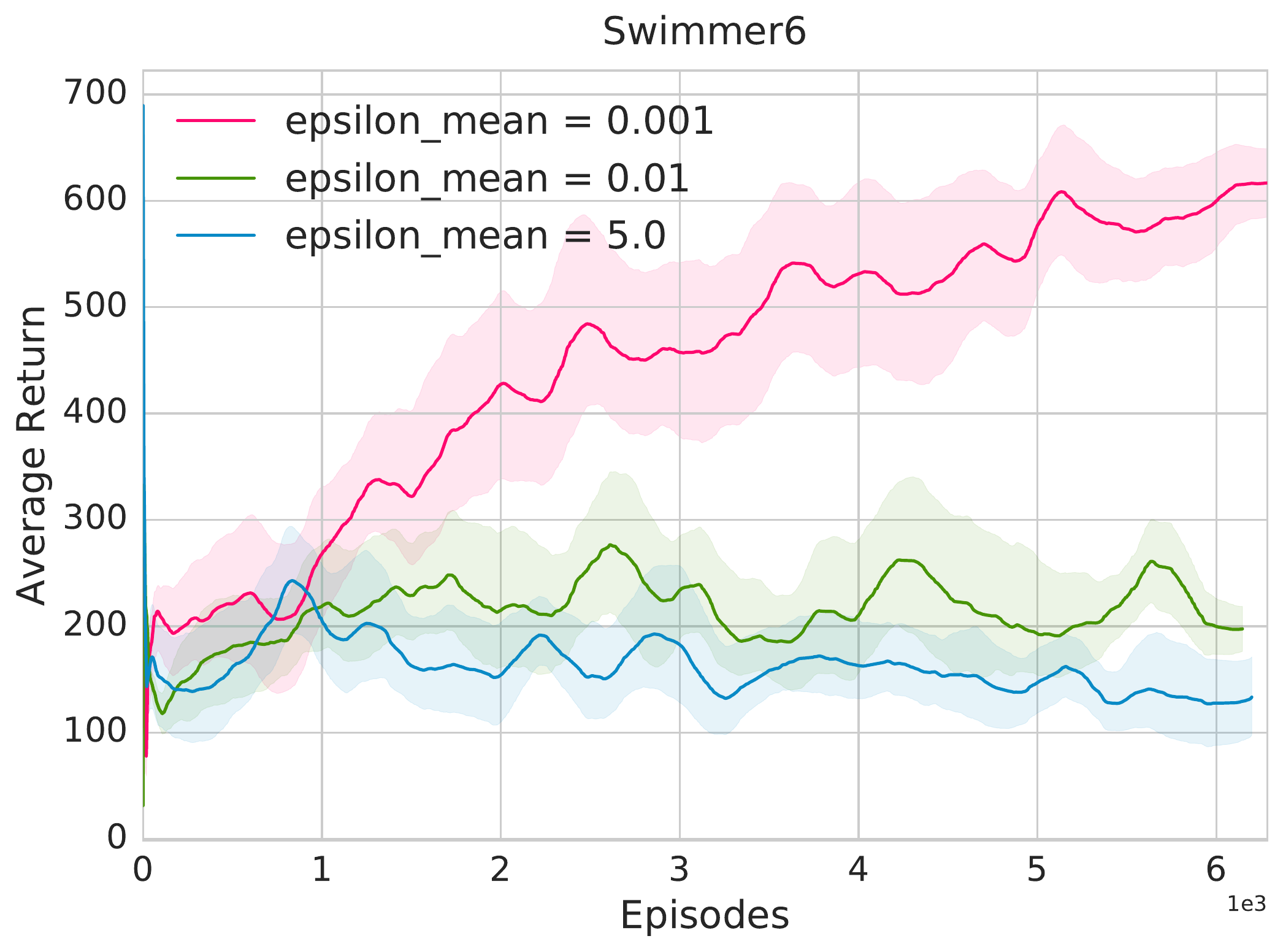}
  \caption{Sweep mean bound}
  \label{fig:meanCovBound:mean}
\end{subfigure}%
\begin{subfigure}{\mywidth\textwidth}
  \centering
  \includegraphics[width=\textwidth]{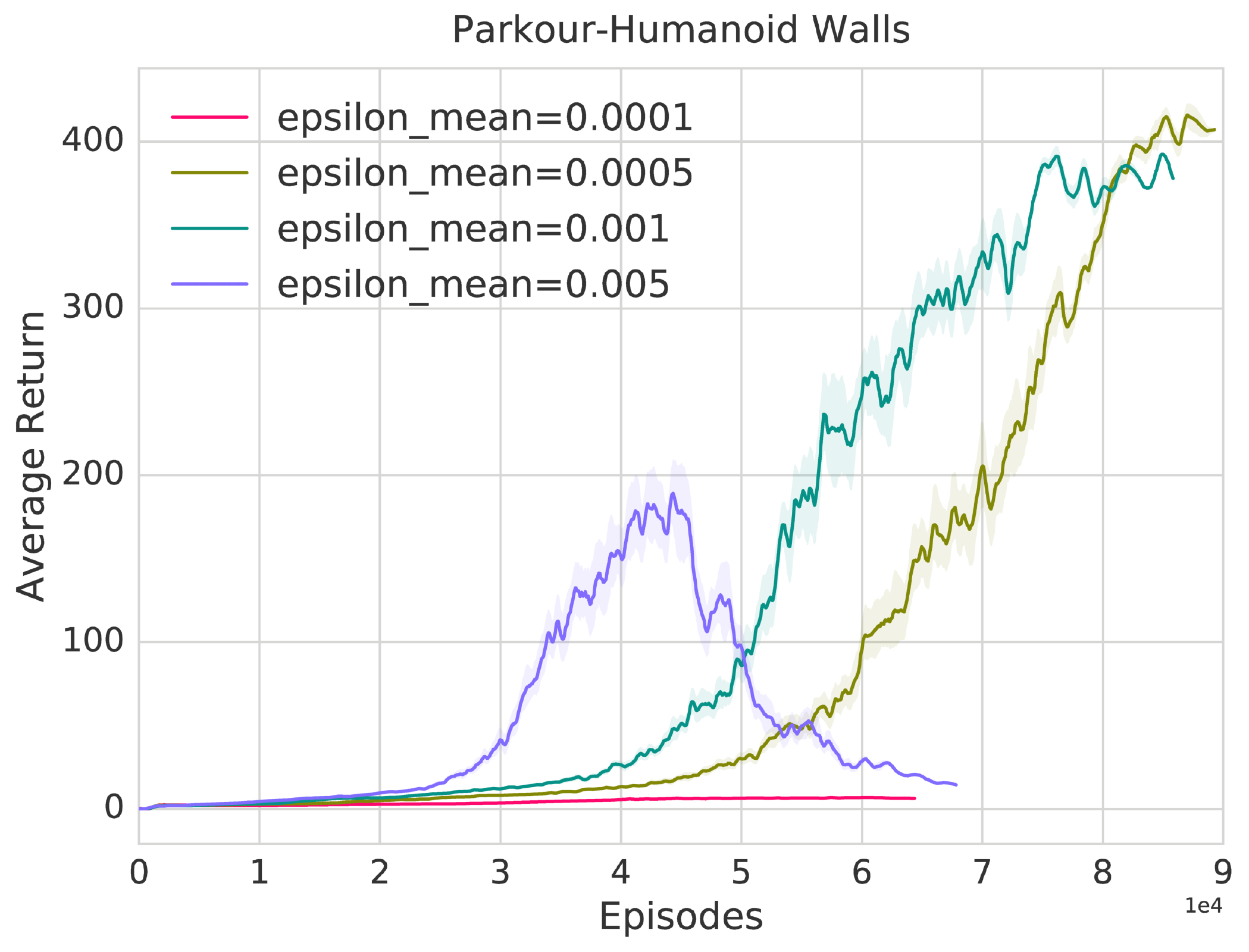}
  \caption{Sweep mean bound}
  \label{fig:meanCovBound:mean}
\end{subfigure}%
\begin{subfigure}{\mywidth\textwidth}
  \centering
  \includegraphics[width=\textwidth]{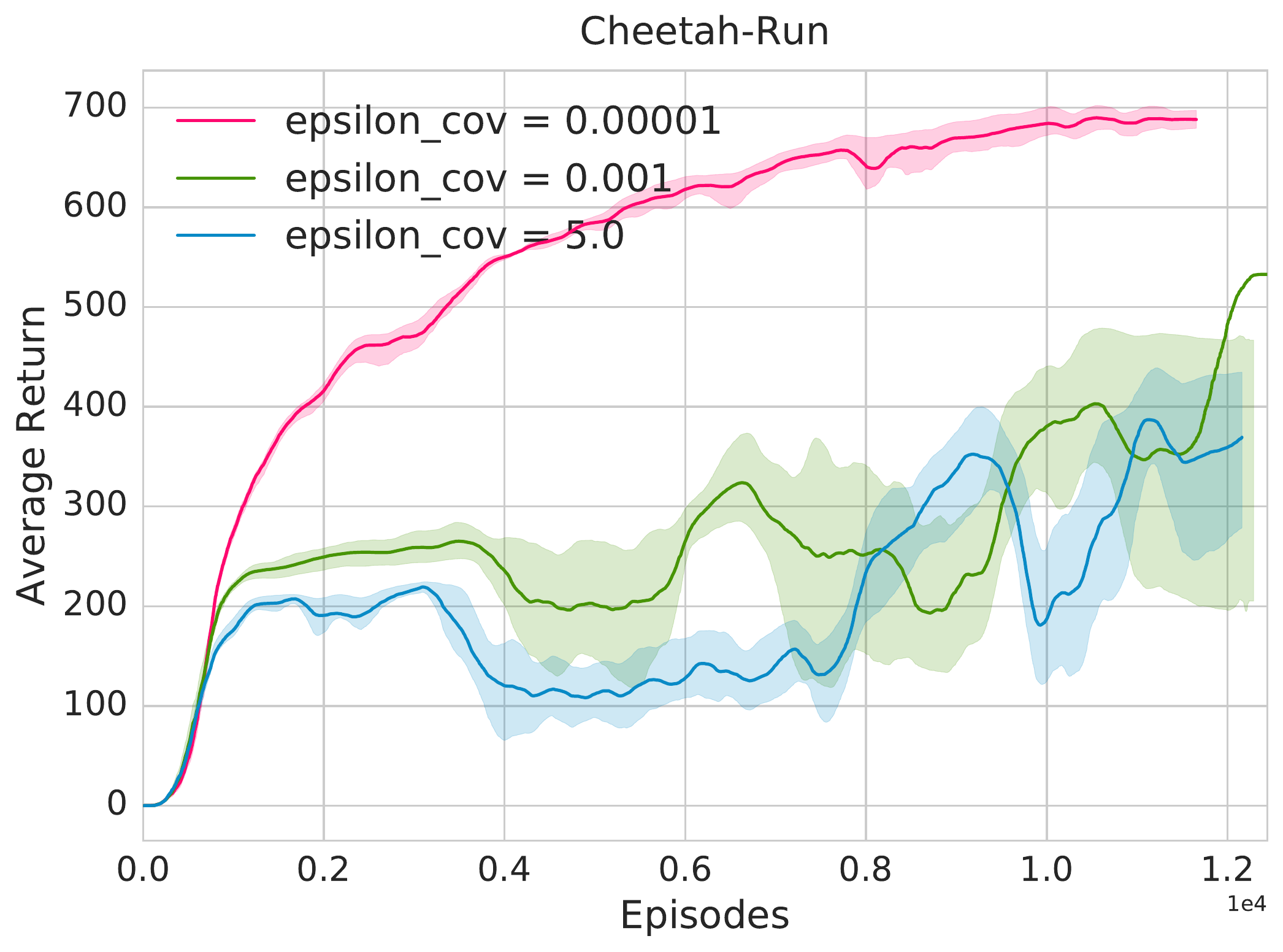}
  \caption{Sweep cov. bound}
  \label{fig:meanCovBound:cov}
\end{subfigure}
\caption{(a): We evaluate the effect of using a KL bound for the decoupled MLE update in the humanoid-stand task.  (b): We fix the KL bound on the covariance and evaluate three different bounds for the mean. The results clearly show that the most conservative bound on the change of the mean(0.001) leads to the best results. (c): Same as (b) but on the parkour-walls task. Here, we see that a loose bound on the mean decreases performance and a too tight bound slows down learning (d): we fix the KL bound on the mean and evaluate different bounds on the covariance (in the cheetah domain). Overall the results emphasize the importance of conservative updates.}
\label{fig:meanCovBound}
\vspace{-0.4cm}
\end{figure*}
\begin{figure*}[ht]
\centering
\begin{minipage}[c]{1\textwidth}
\def\mywidth{0.24}
\def\myhsep{-0.01}
\includegraphics[width=\mywidth\textwidth]{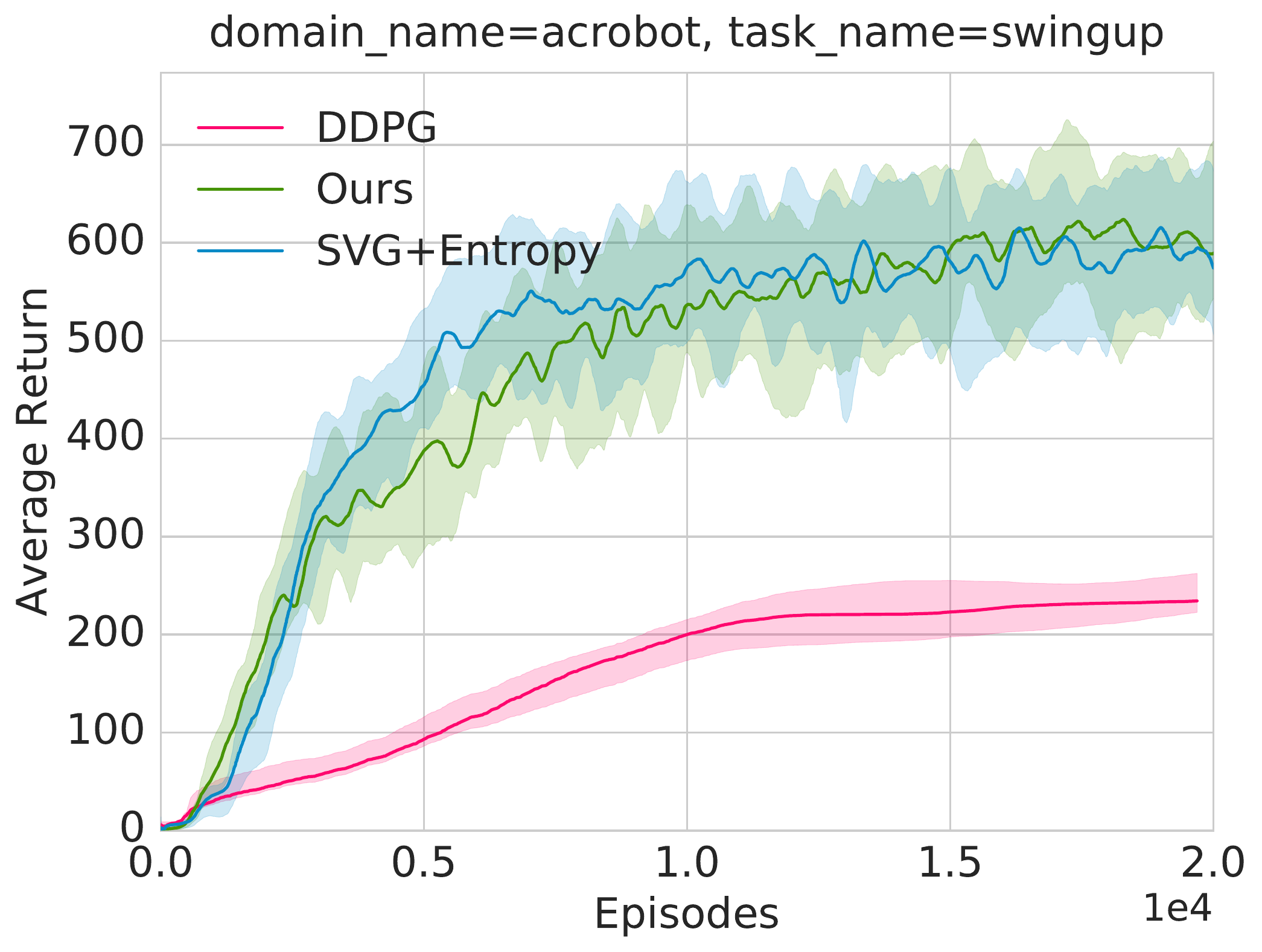}
\hspace{\myhsep\textwidth}
\includegraphics[width=\mywidth\textwidth]{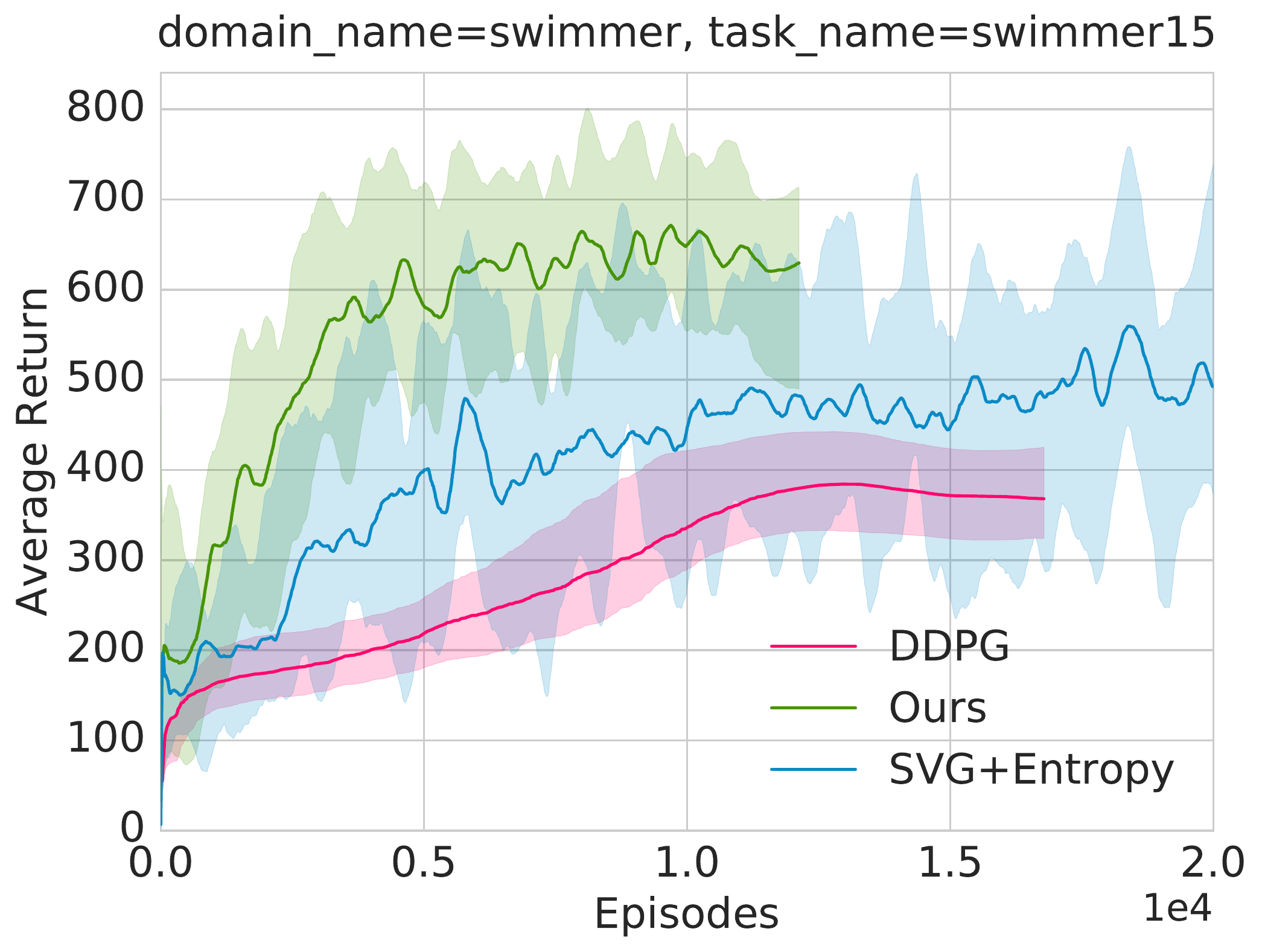}
\hspace{\myhsep\textwidth}
\includegraphics[width=\mywidth\textwidth]{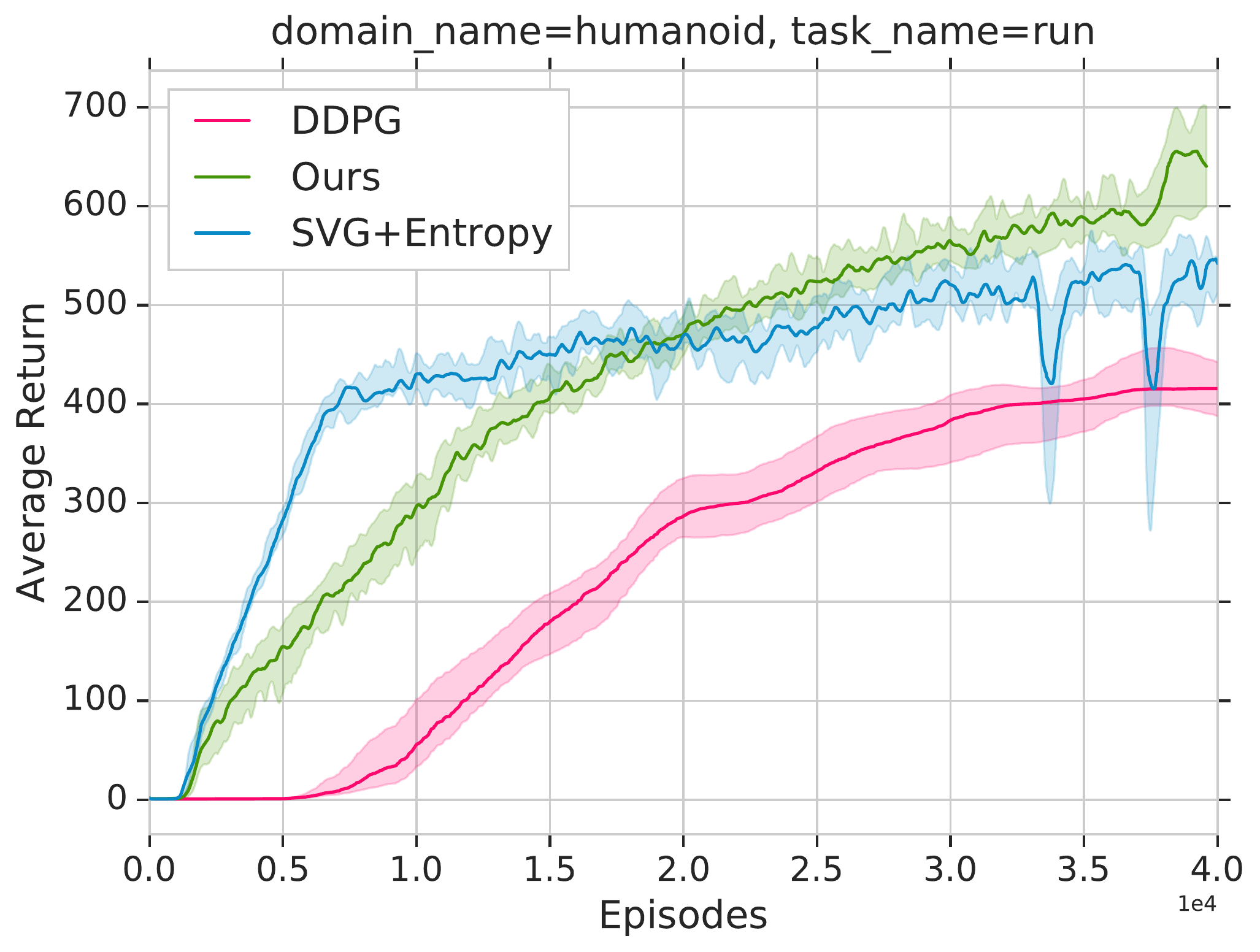}
\hspace{\myhsep\textwidth}
\includegraphics[width=\mywidth\textwidth]{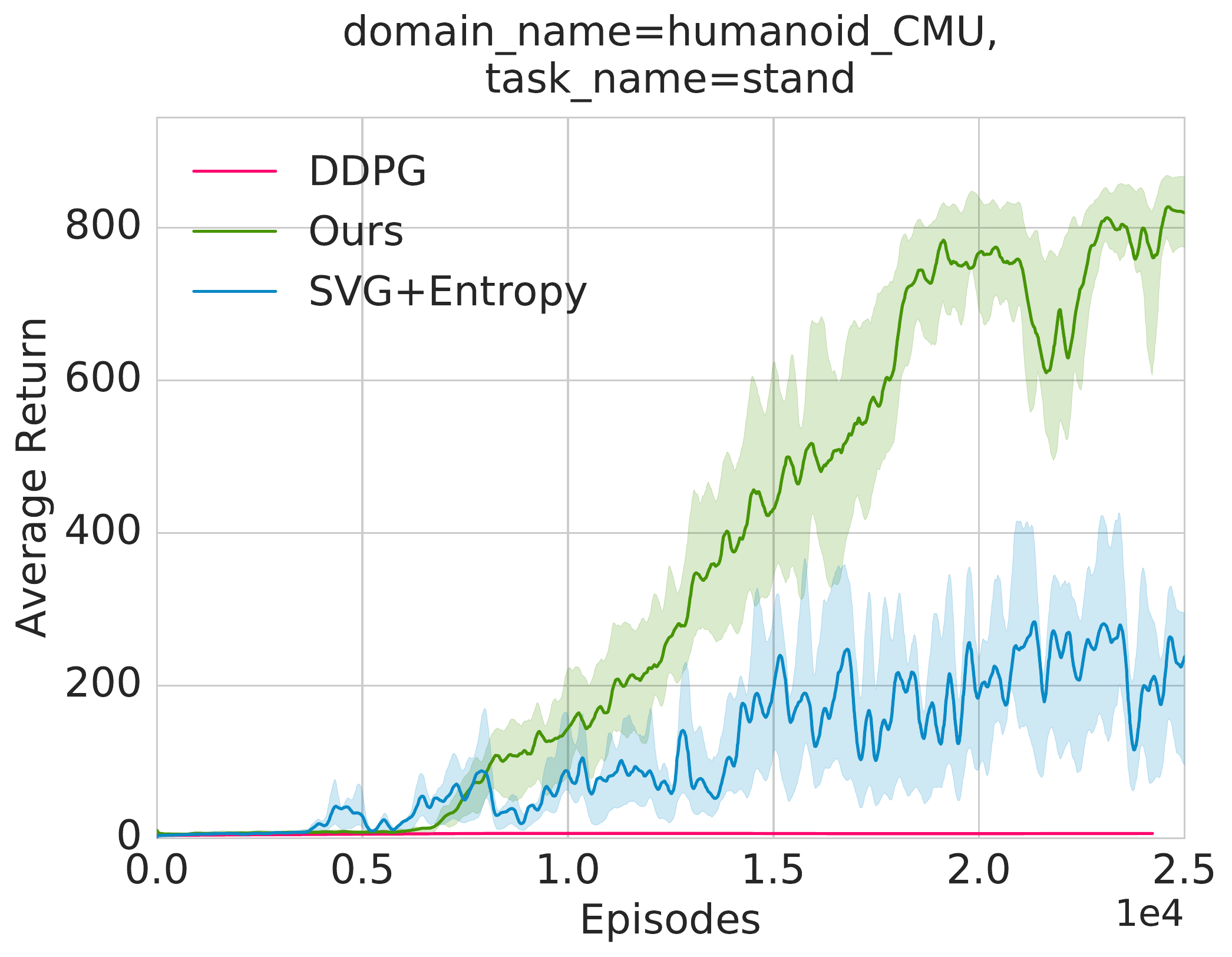}

\caption{Comparison between our algorithm, SVG and DDPG. Results show that while all algorithms perform similar in low-dimensional tasks when hyperparameters are tuned correctly; differences start to emerge in high dimensional tasks. I.e. in the swimmer with 15 links, humanoid run and humanoid CMU-stand tasks where our algorithm performs more stable and achieves better performance.}
\label{fig:svg-mpo}
\end{minipage}
\vspace{-0.3cm}
\end{figure*}

\begin{figure*}[ht]
\centering
\def\mywidth{0.24}
\begin{subfigure}{\mywidth\textwidth}
  \centering
  \includegraphics[width=\textwidth]{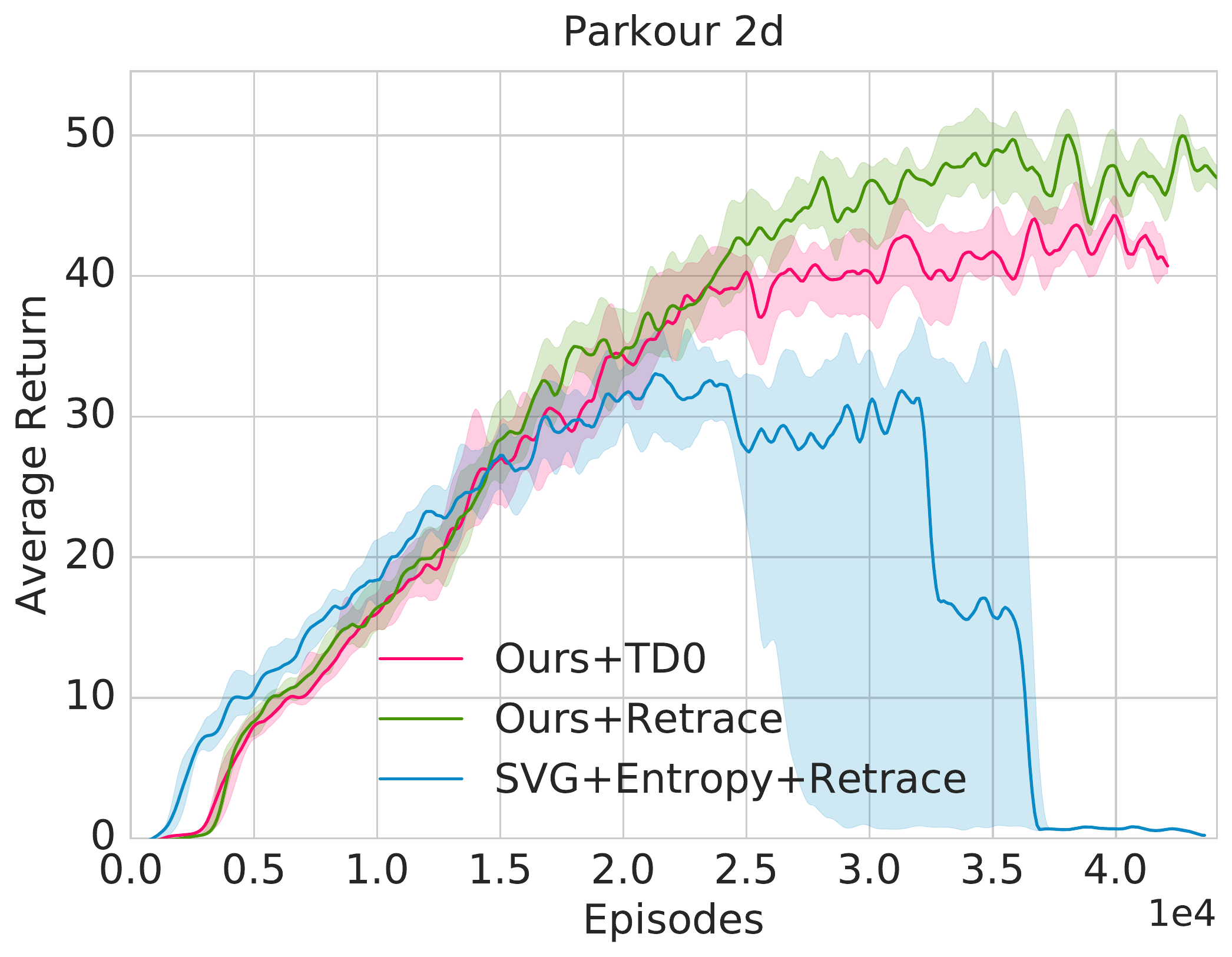}
  \caption{Parkour-2Dwalker}
  \label{fig:sub1}
\end{subfigure}%
\begin{subfigure}{\mywidth\textwidth}
  \centering
  \includegraphics[width=\textwidth]{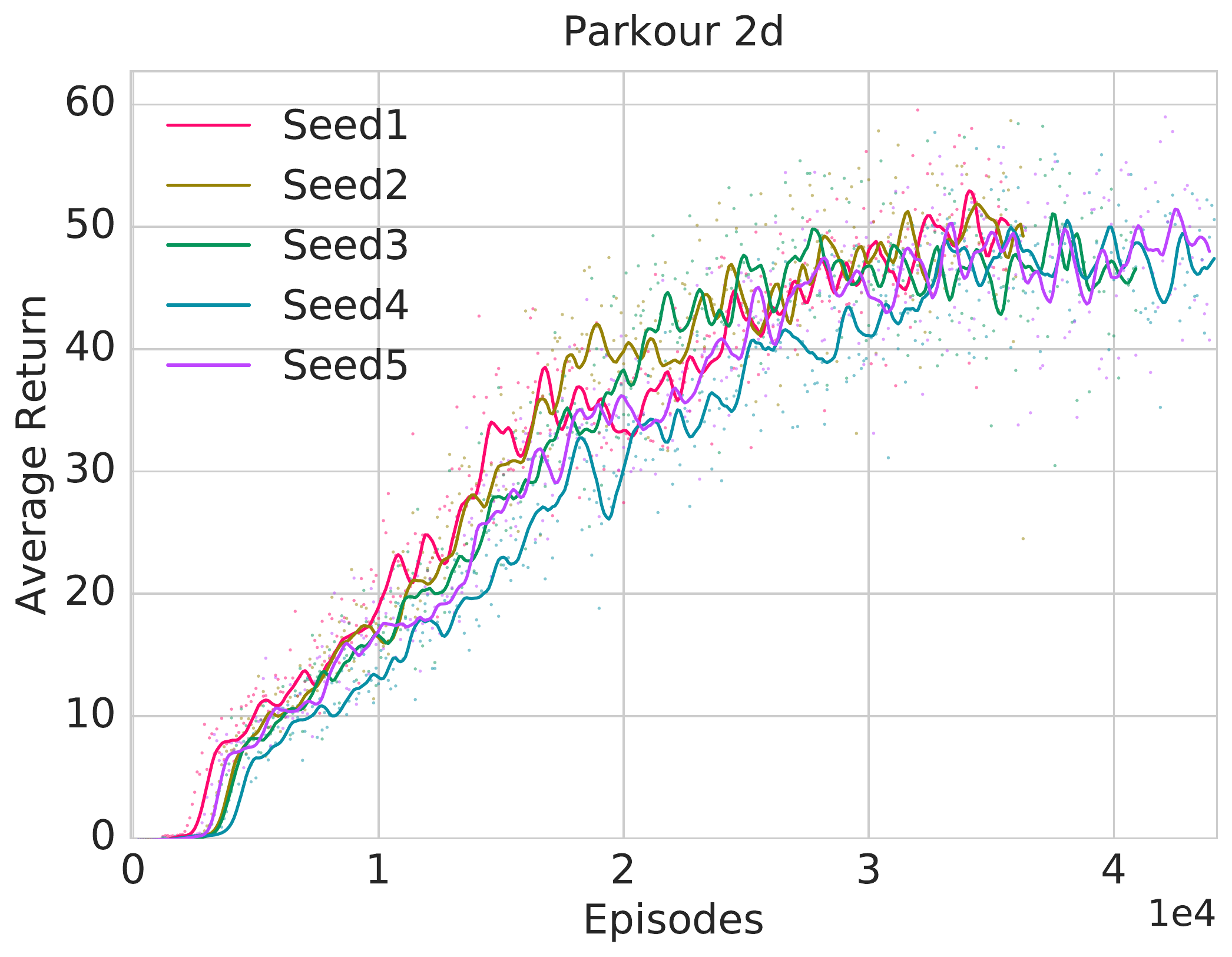}
  \caption{Parkour2D seeds}
  \label{fig:sub1}
\end{subfigure}%
\begin{subfigure}{\mywidth\textwidth}
  \centering
  \includegraphics[width=\textwidth]{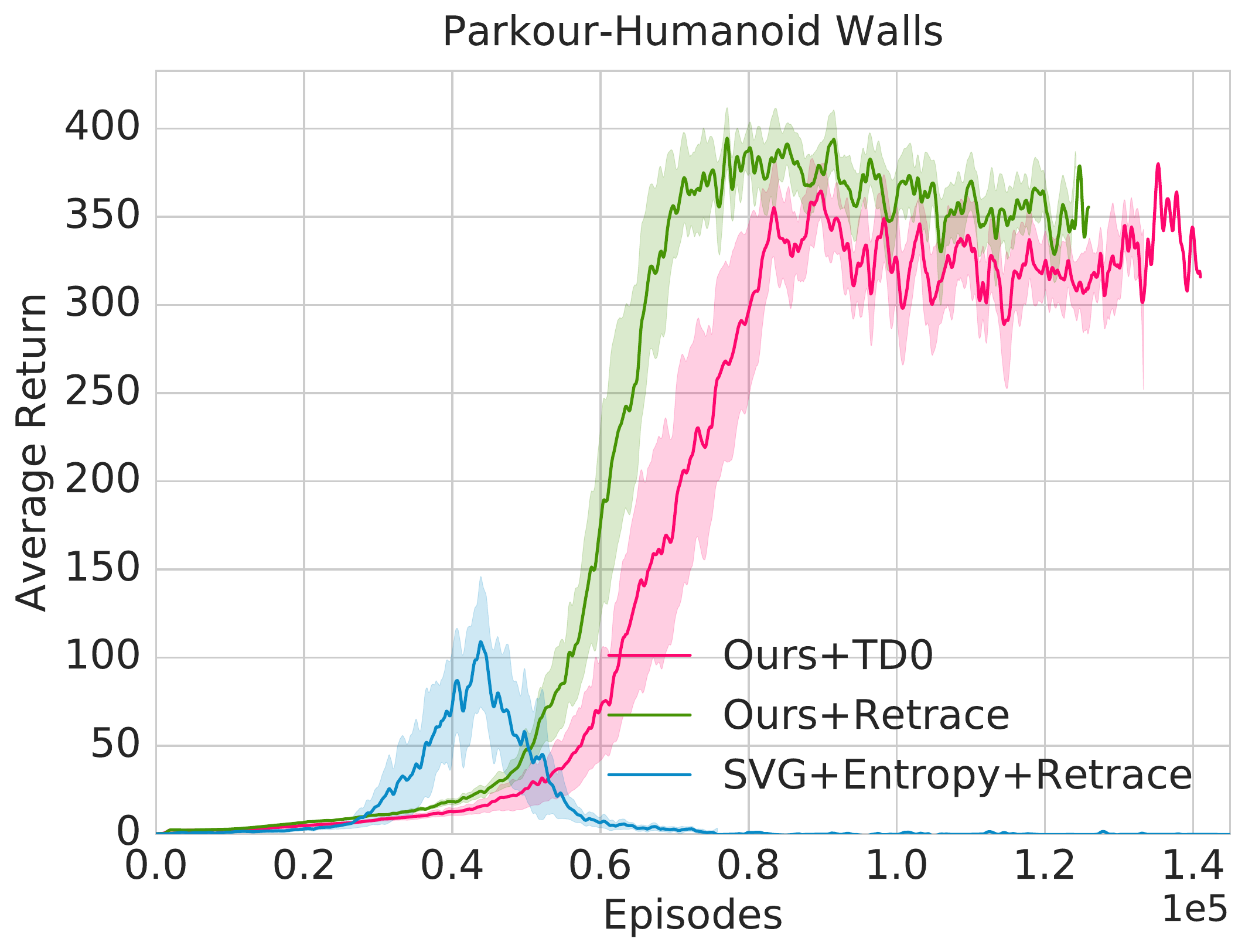}
  \caption{Humanoid walls}
  \label{fig:sub2}
\end{subfigure}
\begin{subfigure}{\mywidth\textwidth}
  \centering
  \includegraphics[width=\textwidth]{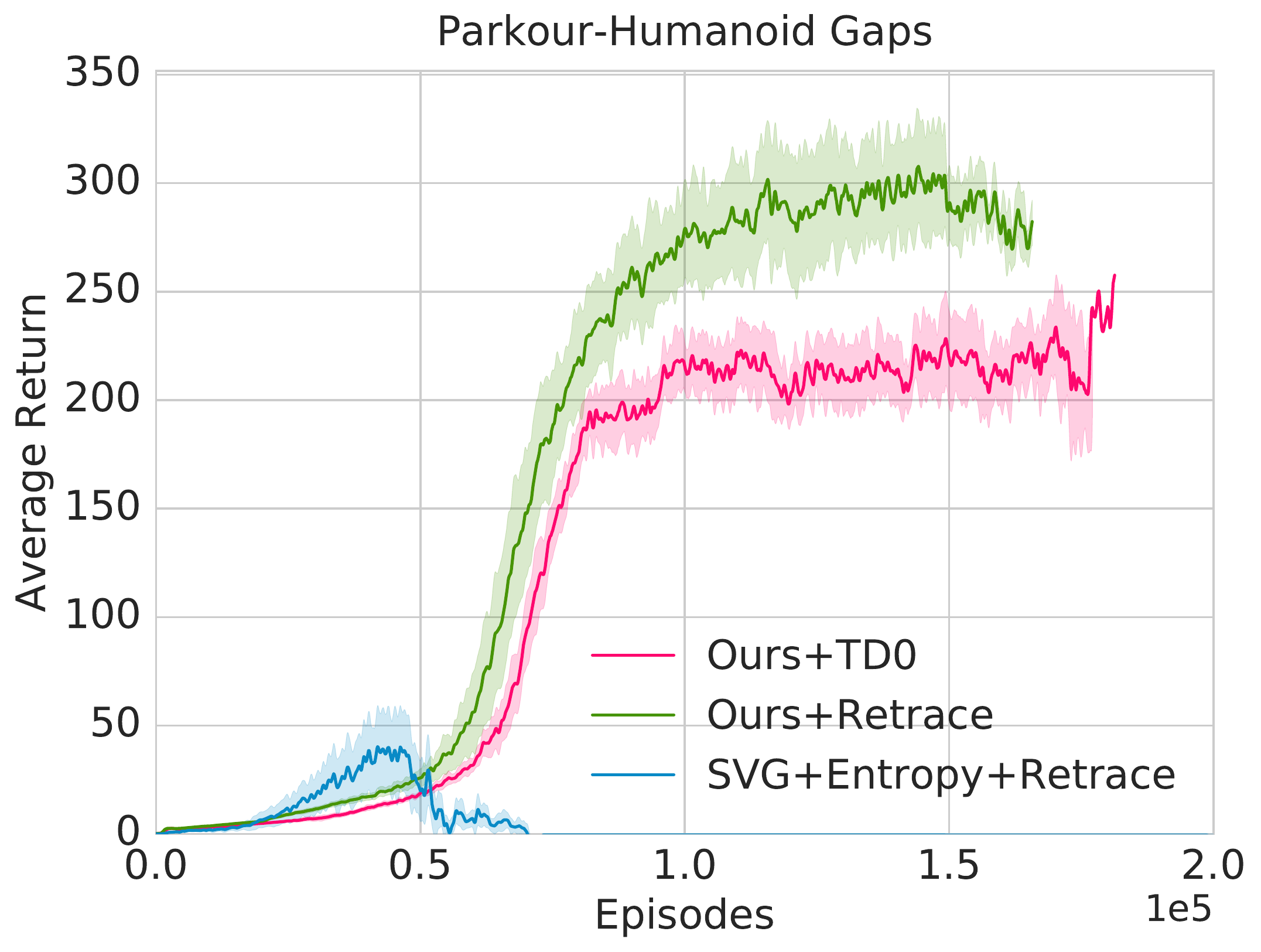}
  \caption{Humanoid gaps}
  \label{fig:sub2}
\end{subfigure}
\caption{Evaluation on high dimensional parkour tasks as illustrated in Figure 10 in the appendix. We use our algorithm with two different policy evaluation methods: TD0 and Retrace. We also plot SVG as a baseline. We are able to learn all three tasks over all 5 seeds but see a benefit for retrace in difficult tasks, as it converges faster with better asymptotic performance.}
\label{fig:parkour-tasks}
\vspace{-0.5cm}
\end{figure*}


\subsubsection{Experimental setup}
Unless noted otherwise we use the decoupled updates proposed in Section \ref{sec:gaussian} in combination with the exponential transformation. Experiments with ranking based weights are given in the appendix in Section D. 
The policy is a state-conditional Gaussian parameterized by a feed-forward neural network. 
We use a single learner process on a GPU and a single actor process on a CPU to gather data from the environment, performing asynchronous learning from a replay buffer. Unlike e.g.\ in \citet{d4pg} we do not perform distributed data collection. We use a single fixed set of hyperparameters across all tasks to show the reliability of the proposed algorithm. Details on all network architectures used, and the hyperparameters are given in the appendix in section G.
For each task, we repeat the experiments five times and report the mean performance and standard deviation. 

\subsubsection{Control Suite Tasks}

Full results are given in Figure 11 and Figure 12 in the appendix. We here focus on five domains for detailed comparisons: the acrobot (with 2 action dimensions), the swimmer (15 action dimensions), the cheetah (6 action dimensions), the humanoid (22 action dimensions) and the CMU humanoid (56 action dimension), as illustrated in the appendix in Figure 9.

\paragraph{Ablations}
We consider four ablations of our algorithm comparing: i) the full algorithm, ii) no KL constraints ii) varying the strength of the KL on the mean, iii) varying the strength of the KL on the covariance.
First, we compare the optimization with KL bounds on the mean and the covariance with a variant when there is no KL bound. I.e. fitting is performed via MLE. As depicted in Figure \ref{fig:meanCovBound:kl}, without a constraint learning becomes unstable and considerably lower asymptotic performance is achieved.
We found that decoupling the policy improvement objective for the mean and covariance can alleviate premature convergence. In Figures~\ref{fig:meanCovBound:mean} and Figure~\ref{fig:meanCovBound:cov} we compare for two environments different settings for the bounds on the mean while keeping the bound on the covariance fixed to the best value obtained via a grid-search, and vice versa.
The results show that bounding both mean and covariance matrix is important to achieve stable performance. 
This is consistent with  existing studies which have previously found that avoiding premature convergence (typically by tuning exploration noise) can be vitally important \cite{duan2015benchmarking}. In general, we find that constraints are important for reliable learning across tasks.

\paragraph{Comparison to DDPG and SVG}
Figure~\ref{fig:svg-mpo}, shows comparisons to two optimized reference implementations of DDPG~\cite{lillicrap2015continuous} and SVG(0)~\cite{heess2016learning}, using the same asynchronous actor-learner separation and Q-learning procedure as for our algorithm. For both baselines we place a tanh on the mean of the Gaussian behaviour policy. While DDPG uses a fixed variance, for SVG we also set a minimum variance by adding a constant of $0.1$ to the diagonal covariance. We found this to be necessary for stabilizing the baselines. No restrictions are placed on either mean or covariance in our method. For SVG we used entropy regularization with a fixed coefficient (we refer to Section G.2 in the appendix for details). All algorithms perform similar in the low-dimensional tasks, but differences emerge in higher-dimensional tasks. Overall, when using a single set of hyperparameters for all tasks, our algorithm is more stable than the reference algorithms. Especially in problems with a high dimensional action space it achieves a better asymptotic performance than the baselines.\footnote{We note that better performance could be obtained by tuning on a per-task basis for all algorithms.} 

\subsubsection{Parkour tasks}
In this section, we consider three Parkour tasks \citep{heess2017emergence}. These tasks require the policy to steer a robotic body through an obstacle course, performing jumps and avoidance maneuvers. A depiction of the environments can be found in the appendix, Figure 10. We use the same setup and hyperparameters as in the previous section. This includes still using only one GPU for learning and one actor for interacting with the environment, a significant reduction in compute compared to the previously used 32-128 actors for solving these tasks~\cite{abdolmaleki2018maximum,d4pg,heess2017emergence}.
We compare two variants of our method with two variants of SVG and DDPG: a version with TD0 to fit the Q-function, and a version with Retrace~\citep{munos2016safe}. 

As shown in Figure~\ref{fig:parkour-tasks}, only our method is able to solve all tasks. In addition, \emph{our algorithm is capable of solving these challenging tasks while running on a single workstation}. Analyzing the results in more detail we can observe that learning the Q-function with TD(0) leads to overall slower learning than when using Retrace, as well as to lower asymptotic performance. This gap increases with task complexity. Among the parkour tasks, \emph{humanoid-3D gaps} is the hardest as it requires controlling a humanoid body to jump across gaps, see Figure 10 in the appendix. \emph{Parkour-2D} is the easiest and only requires the policy to control a walker in a 2D environment. We observed a similar trend for other tasks although the difference is less dramatic on low-dimensional tasks such as the ones in the control suite. 
\begin{figure*}[t]
\centering
\begin{minipage}[c]{1\textwidth}
\def\mywidth{0.24}
\def\myhsep{-0.01}
\includegraphics[width=\mywidth\textwidth]{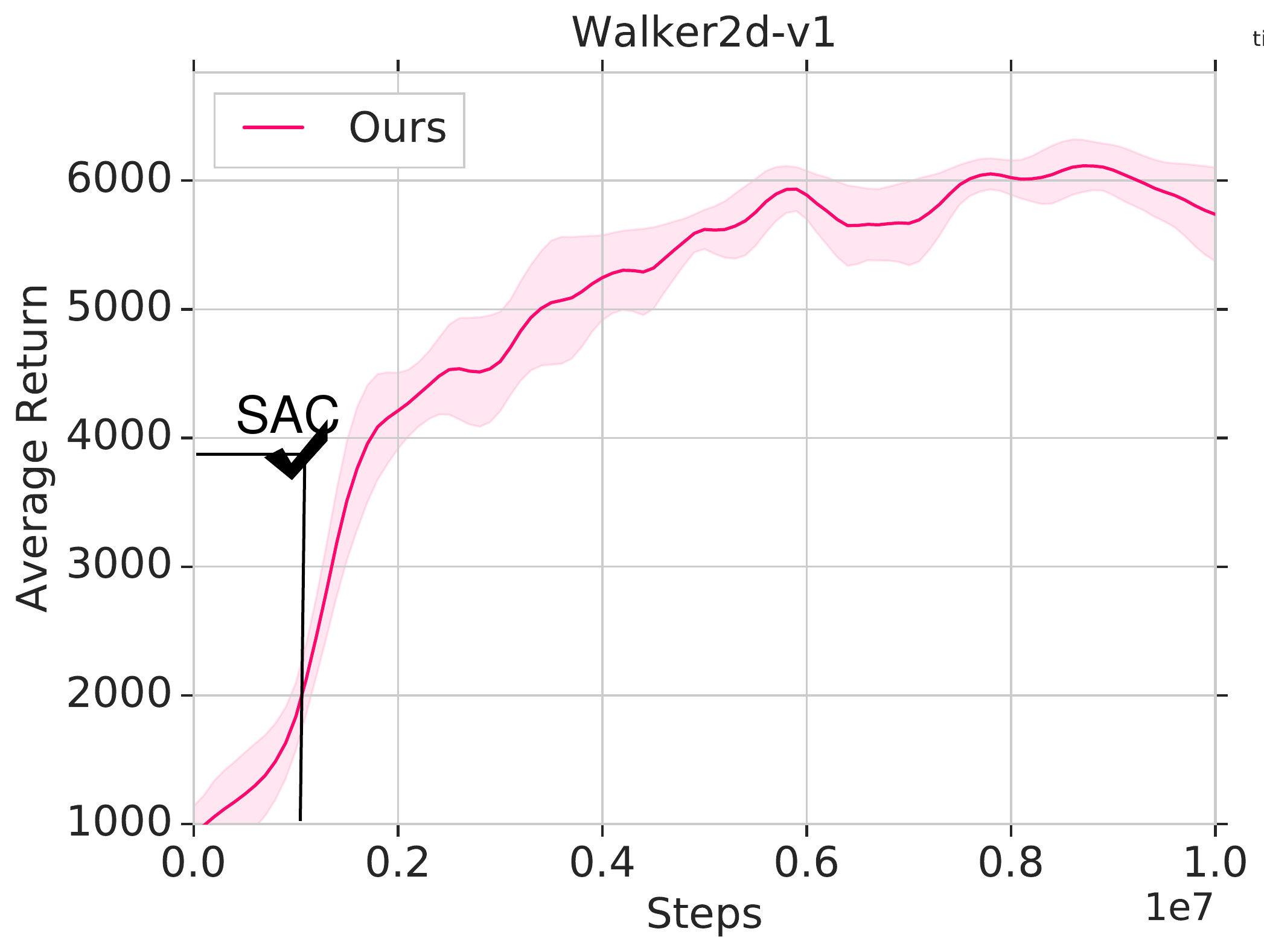}
\hspace{\myhsep\textwidth}
\includegraphics[width=\mywidth\textwidth]{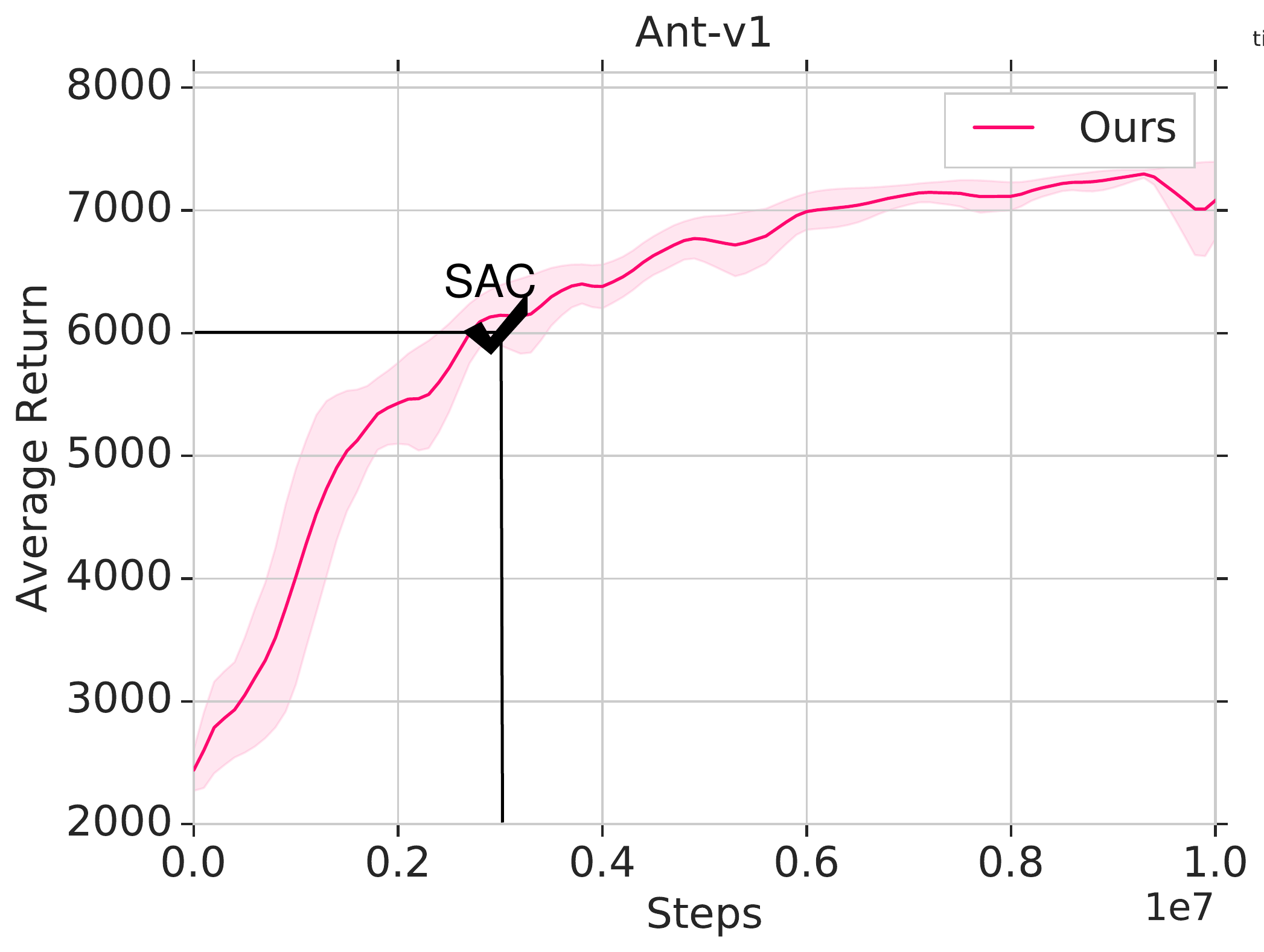}
\hspace{\myhsep\textwidth}
\includegraphics[width=\mywidth\textwidth]{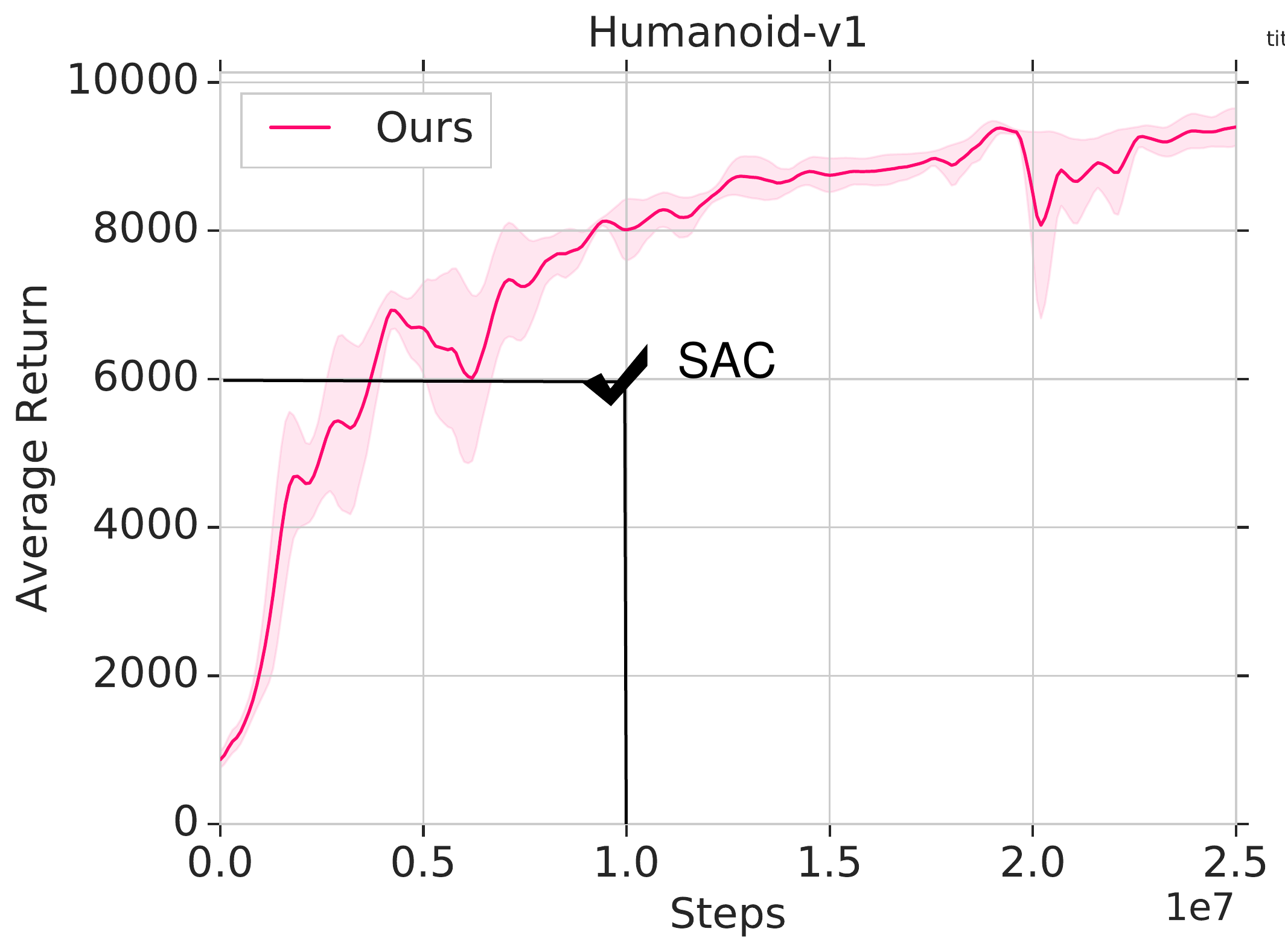}
\hspace{\myhsep\textwidth}
\includegraphics[width=\mywidth\textwidth]{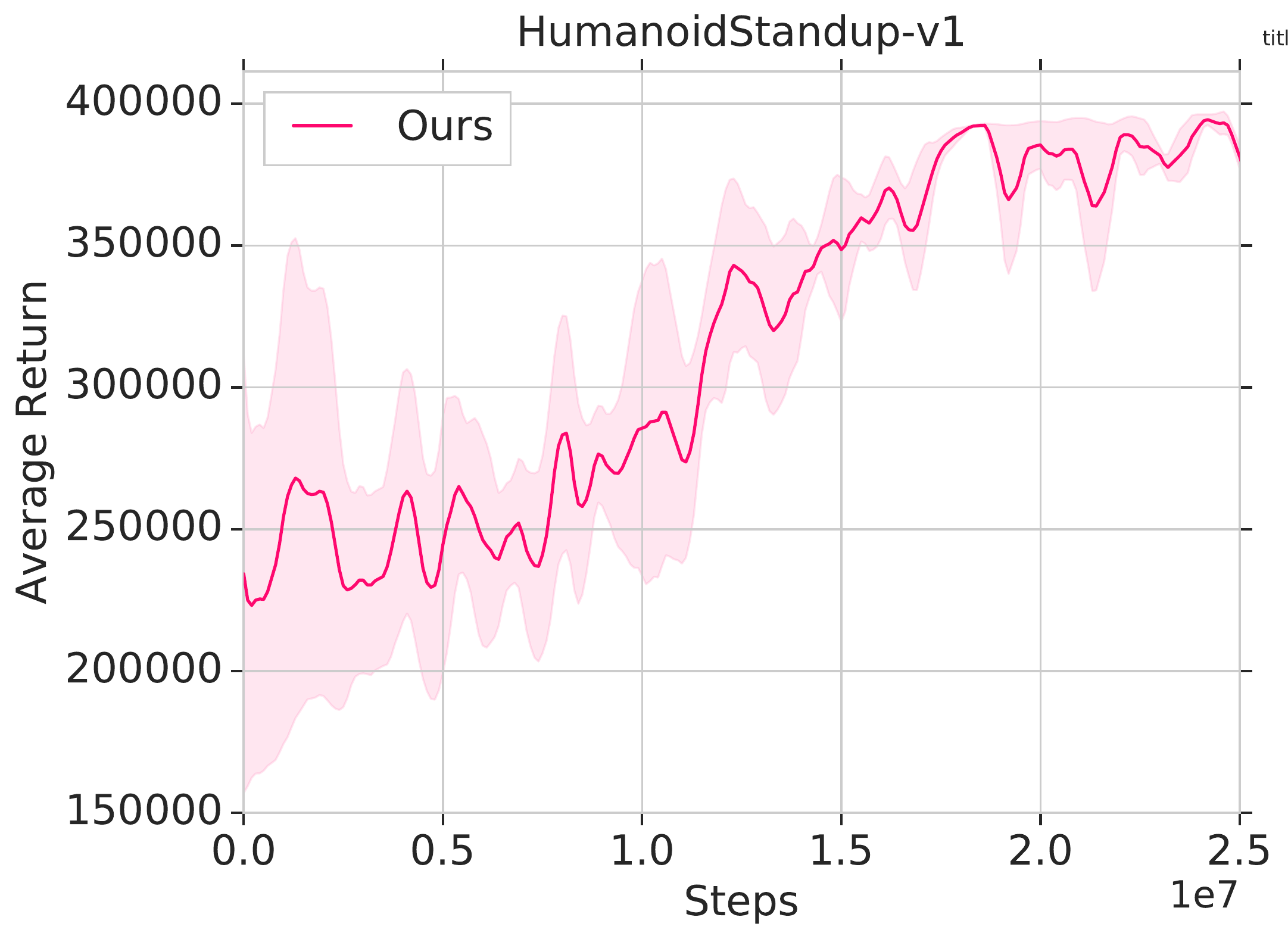}

\caption{Comparison between our algorithm and SAC on walker, ant and humanoid from OpenAI gym. Check-mark shows the best reported performance of SAC \citep{haarnoja2018soft}. Results show that our method solve the tasks with same hyper parameters as before while achieving considerably better asymptotic performance than SAC with on-par sample efficiency. SAC does not report any result on humanoid-standup. Humanoid-standup is in particular interesting because of its very different reward scale with respect to other environments. Note that our method can also solve humanoid-stand with final return of 4000000 using the same hyper parameters. }
\label{fig:sac-mpo}
\end{minipage}
\vspace{-0.3cm}
\end{figure*}
\subsubsection{OpenAI Gym}
Finally, we consider OpenAI gym tasks\citep{brochmanopenaigym} to compare our method with soft-actor critic algorithm (SAC)\citep{haarnoja2018soft} which is an actor-critic algorithm very similar to SVG(0) that optimizes the entropy regularized objective expected reward objective. We use four tasks from OpenAI gym \citep{brochmanopenaigym}, i.e, Ant, Walker2d, Humanoid run, Humanoid stand-up for evaluating our method against SAC. For policy evaluation we use Retrace~\citep{munos2016safe}. We report the evaluation performance as in \cite{haarnoja2018soft} every 1000 environment steps and compare to their final performance. To obtain  a similar data generation rate to \cite{haarnoja2018soft} we slowed down the actor such that it generated 1 trajectory each 5 seconds. We used the same hyperparameters for learning as we used for Parkour suite and DeepMind control suite. Our results in figure \ref{fig:sac-mpo} show that we achieve considerably better asymptotic-performance than the ones reported by SAC\citep{haarnoja2018soft} in these environments with on-par sample efficiency. With thesame hyper parameters, our method also solves humanoid-stand with final return of 4000000 which is 100-1000 order of magnitude different than the final return of other environments. 

\section{Conclusion}
We have presented a policy iteration algorithm for high-dimensional continuous control problems. The algorithm alternates between Q-value estimation, local policy improvement and parametric policy fitting; hard constraints control the rate of change of the policy. And a decoupled update for mean and covarinace of a Gaussian policy avoids premature convergence. Our analysis shows that when an approximate Q-function is used, slow updates to the policy can be critical to achieve reliable learning. Our comparison on 31 continuous control tasks with rather diverse properties using a limited amount of compute and a single set of hyperparameters demonstrate the robustness our method while it achieves state of art results. 
\balance
\bibliographystyle{abbrv}
\bibliography{references}

\appendix

\section{Relation to Policy Gradients}
An interesting possibility is to use an identity transformation in Step 2 of our alogrithm (instead of using ranking or an exponential transformation). While not respecting the desiderata i) and ii) from above this would bring our method close to an expected policy gradient algorithm \citep{EPGCiosek}. We discuss this choice in detail in the appendix.
It is instructive to also consider the case of an identity $w_{ij} = id(Q^{\pi^{(k)}}(a_i, s_j)) = Q^{\pi^{(k)}}(a_i, s_j)$ as the transformation function. Clearly, the identity is rank preserving. It does, however, not satisfy the additional requirements i) positivity of weights, and ii) weights are normalized such that $\sum_i w_{ij} = 1$ outlined in the main paper (weights can be negative and are not normalized). This hints at the fact that it would make our procedure susceptible to instabilities caused by scaling Q-values and can result in ``agressive'' changes of the policy distribution away from bad samples (for which we have negative weights). Considering the identity is, nonetheless, an interesting exercise as it highlights a connection of our algorithm to a likelihood-ratio policy gradient \citep{SuttonPGM} approach since we would obtain: $\max_{\pi_\a} \sum_j^K \sum_i^N Q^{\pi^{(k)}}(a_{ij}, s_j) \log \pi_\a(a_{ij} | s_j) \approx \max_{\pi_\a} \sum_j^K \mathbb{E}_{\pi^{(k)}(\cdot, | s_j)} \Big[ Q^{\pi^{(k)}}(a, s) \log \pi_\a(a, | s_j) \Big]$; which looks similar to the expected policy gradient (EPG) \citep{EPGCiosek}, where multiple actions are also used to estimate the expectation. On closer inspection, however, one can observe that the above expectation is w.r.t. samples from the old policy $\pi^{(k)}$ and not w.r.t. $\pi_\theta$ (which would be required for EPG). Equivalence to a policy gradient can hence only be achieved for the first gradient step (for which $\pi^{(k)} = \pi_\theta$).



\section{Policy Improvement as Inference}
In the paper, we motivated the policy update rules from a more intuitive perspective. In this section we use inference to derive our policy improvement algorithm. The E-and M-step that we derive here, directly correspond to Step 2 and 3 in the main paper. First, we assume that the Q-function is given and we would like to improve our policy given this Q-function, see algorithm \ref{alg:MPONON} in the paper. In order to interpret this goal in a mathematical sense, we assume there is an observable binary improvement event $R$. When $R=1$, our policy improved and we have achieved our goal. Now we ask, if the policy would have improved, i.e. $R=1$, what would the parameters $\a$ of that improved policy be? More concretely, we can optimize the maximum a posteriori $p(\a|R=1)$ or equivalently:

$$
\argmax_\a \log p(\a|R=1),
$$

after marginalizing out action and state and considering random variable dependencies, it is equivalent to optimizing $$\argmax_\a \log\int\mu(s)\int p(R=1|a,s)\pi(a|s,\a)p(\a) \diff a \diff s $$

Here $\mu(s)$ is the stationary state distribution and is given in each policy improvement step. In our case, $\mu(s)$ is the distribution of the states in the replay buffer. $\pi(a|s,\a)$ is the policy parametrized by $\a$ and $p(\a)$ is a prior distribution over the parameters $\a$. This prior is fixed during the policy improvement step and we set it such that we stay close to the old policy during each policy improvement step. $p(R=1|a,s)$ is the probability density of the improvement event, if our policy would choose the action $a$ in the state $s$. In the other word $p(R=1|a,s)$ defines the probability that in state $s$, taking action $a$ over other possible actions, would improve the policy. 
As we prefer actions with higher Q-values, this probability density function can be defined by $p(R=1|a,s) = \frac{C(Q(a,s))}{\int C(Q(a,s)) \diff a}$, where $C$ is a monotonically increasing and therefore rank preserving function of Q function. This is a sensible choice, as choosing an action with higher Q-values should have a higher probability of improving the policy in that state.

However, explicitly solving this equation for $\a$ is hard. Yet, the expectation-maximisation algorithm does give an efficient method for maximizing $\log p(\a|R=1)$ in this setting. Therefore, our strategy is to repeatedly construct a lower-bound on this probability density in the E-step, and then optimize that lower-bound in the M-step. Following prior work in the field, we construct a lower bound on $\log p(\a|R=1)$ using the following decomposition,  

\begin{equation*}
\begin{aligned}
\log p(\a | R=1) = \KL(q(a,s)\,\|\,p(a,s|R=1,\a))
+ \iint q(a,s)\log \frac{p(R=1|a,s)\pi(a|s,\a)\mu(s)p(\a)}{q(a,s)} \diff a \diff s
\end{aligned}
\end{equation*}
where $q(a,s)$ is an arbitrary variational distribution. Please note that the second term is a lower bound as the first term is always positive. In effect, $\pi(a|s,\a)$ and $q(a|s) = \frac{q(a,s)}{\mu(s)}$ are unknown, even though $\mu(s)$ is given. 

We can now focus on the underlying meaning of $q(a|s)$. If in each state we knew which action would lead to a policy improvement, we would fit a policy which outputs that action in each state. However, we do not have access to that knowledge. Instead, in the E-step we use the Q-function to infer a distribution over the actions $q(a|s)$ of which we know that choosing those actions would improve the policy. In the M-step, we then fit a policy to this distribution $q(a|s)$ such that those actions are selected by the newly fitted policy, hence the policy is improved.

\subsection{E-Step (Step 2 in main paper)}
In the E-step (which would correspond to Step 2 in the main paper), we choose the based variational distribution $q(a|s) = \frac{q(a,s)}{\mu(s)}$ (approximated via the sample based distribution in the main paper) such that the lower bound on $\log p(\a_t|R=1)$ is as tight as possible. We know this is the case when the KL term is zero given the old policy $\a_t$. Therefore we minimize the KL term given the old policy, i.e,
\begin{align*}q(a|s) = \argmin_{q} \int \mu(s) \KL(q(a|s)\,\|\,p(a,s|R=1,\a_t) \diff s \end{align*} 

which is equivalent to minimizing, 
\begin{align} 
q(a|s) = \argmin_{q} \int \mu(s) \KL(q(a|s)\,\|\,\pi(a|s,\a_t) )\diff s 
- \int \mu(s)\int q(a|s) \log p(R=1|a,s))\diff a\diff s
\label{eq:estep}
\end{align}

We can solve this optimization problem in closed form, which gives us

$$q(a|s) = \frac{\pi(a|s,\a_t)p(R=1|a,s)}{\int \pi(a|s,\a_t)p(R=1|a,s) \diff a}.$$

Please note that here we only solve for $q(a|s)$ as the state distribution $\mu(s)$ is given and should remain unchanged.

This solution weighs the actions based on their relative improvement probability $p(R=1|a,s)$. At this point we can define $p(R=1|a,s)$ using any arbitrary positive function $C$. For example, we could rank the actions based on their Q-values and assigning positive values to the actions based on their ranking. Alternatively we could define $p(R=1|a,s) \propto \exp\Big(\frac{Q(a,s)}{\eta}\Big)$. Note that temperature term $\eta$ is used to keep the solutions diverse, as we would like to represent the policy with a distribution of solutions instead of only one single solution. Yet, we imply a preferences over solutions by weighing them. However, tuning the temperature $\eta$ is difficult. In order to optimize $\eta$, we plug the exponential transformation in \eqref{eq:estep} and after rearranging terms, our optimization problem is
\begin{align} 
q(a|s) = \argmax_{q} \int \mu(s)\int q(a|s) Q(s,a) \diff a \diff s - 
\alpha \int \mu(s) \KL(q(a|s) \,\|\, \pi(a|s,\a_t) ) \diff s \nonumber 
\end{align}  
or instead of treating the KL bound as a penalty, we can enforce the bound as a constraint: 
\begin{align}
& q(a|s) = \argmax_{q} \int \mu(s)\int q(a|s) Q(s,a) \diff a \diff s \\ 
& \textrm{s.t.} \int \mu(s) \KL(q(a|s) \,\|\, \pi(a|s,\a_t) ) \diff s < \epsilon. \nonumber 
\end{align}
Note that when the $q(a|s)$ is parametric this is the policy optimization objective for MPO-parametric~\citep{abdolmaleki2018maximum}, TRPO~\citep{schulman15} , PPO~\citep{schulman2017ppo} and SAC~\citep{haarnoja2018soft}(if old policy is a uniform distribution). Note that  
In our case $q(a|s)$ is a non-parametric and samples based distribution, and we can solve this constraint optimization in close form for each sample state $s$,
$$q(a|s) \propto \pi(a|s,\a_t) \exp\Big(\frac{Q(s,a)}{\eta}\Big)$$
and easily optimize for the correct $\eta$ using the convex dual function.
\begin{align*}
\eta=\argmin_{\eta}
\eta\epsilon+\eta\int\mu(s)\log\int \pi(a|s,\a_t)\exp\Big(\frac{Q(s,a)}{\eta}\Big)\diff a \diff s
\end{align*}

Please see section \ref{sec:dualfunctionderivation} for dual function derivation details. Now if we estimate the integrals using state samples from replay buffer and our old policy we recover the policy and dual function given in Step 2 of the main paper.

\subsection{M-step (Step 3 in main paper)}

Since we obtained the variational distribution $q(a|s)$, we have found a tight lower bound to our density function $p(\a|R=1)$. Now we can optimize the parameters $\a$ of the policy $\pi(a|s,\a)$ in order to maximize this lower bound,
\begin{align*}
\a = \argmax_{\a} \int \mu(s) \int q(a|s)\log \frac{\pi(a|s,\a) }{q(a|s)}\diff a \diff s + \log p(\a).
\end{align*}
This corresponds to the maximum likelihood estimation step (Step 3) in the main paper.
 
As a prior $\log p(\a)$ on the parameters $\a$, we can say that the new policy $\pi(a|s,\a)$ should be close to the old policy $\pi(a|s,\a_t)$, or more formal, we can choose \begin{align*}\log p(\a) = -\lambda \int \mu(s) \KL(\pi(a|s,\a_t) \,\|\, \pi(a|s,\a) ) \diff s\end{align*}

Using this approximation, we find a new optimization problem: 
\begin{align*}
\a = \argmax_{\a} \int \mu(s) \int q(a|s)\log \frac{\pi(a|s,\a) }{q(a|s)}\diff a \diff s  -\lambda \int \mu(s) \KL(\pi(a|s,\a_t) \,\|\, \pi(a|s,\a)) \diff s
\end{align*}

Alternatively we can use a hard constraint to obtain:
\begin{align*}
&\a = \argmax_{\a}  \int \mu(s) \int q(a|s)\log \frac{\pi(a|s,\a)}{q(a|s)} \diff a \diff s \\
&\textrm{s.t.} \int \mu(s) \KL(\pi(a|s,\a_t) \,\|\, \pi(a|s,\a) ) \diff s < \epsilon \nonumber 
\end{align*}

Because of the prior, we do not greedily optimize the M-step objective. Therefore our approach belongs to the category of generalized expectation maximization algorithms. Now if we approximate the integrals in the E-step and M-step using the states samples from replay buffer and the action samples from the old policy we will obtain the exact update rules we proposed in paper. Algorithm \ref{Alg:GradientFree} illustrates algorithmic steps.

\subsubsection{Dual function Derivation \label{sec:dualfunctionderivation}}
The E-step with a non-parametric variational distribution solves the following program:

\begin{equation*}
  \begin{aligned}
  & \max_q \int \mu(s)\int q(a|s) Q(s,a) dads\\  
  & s.t. \int \mu(s) \textrm{KL}(q(a|s) , \pi(a|s,\a_t) )da < \epsilon ,\\
  & \iint \mu(s) q(a|s) dads = 1.
  \end{aligned}
\end{equation*}

First we write the Lagrangian equation, i.e,

\begin{equation*}
  \begin{aligned}
L(q,\eta,\gamma) =& \int \mu(s)\int q(a|s) Q(s,a) dads +\\
&\eta\left(\epsilon - \int \mu(s)\int q(a|s)\log \frac{q(a|s)}{\pi(a|s,\a_t)dads}\right) + \gamma\left(1 - \iint \mu(s) q(a|s)dads\right).
  \end{aligned}
\end{equation*}
Next we maximise the Lagrangian $L$ w.r.t the primal variable $q$. The derivative w.r.t $q$ reads,

$${\partial q}L(q,\eta,\gamma) = Q(a,s) - \eta\log q(a|s) + \eta\log \pi(a|s,\a_t) - (\eta-\gamma).$$

Setting it to zero and rearranging terms we get

$$q(a|s) = \pi(a|s,\a_t)\exp\left(\frac{Q(s,a)}{\eta}\right)\exp\left(-\frac{\eta-\gamma}{\eta}\right).$$

However the last exponential term is a normalisation constant for $q$. Therefore we can write,

$$\exp(\frac{\eta-\gamma}{\eta}) = \int \pi(a|s, \a_t)\exp(\frac{Q(s,a)}{\eta})da.$$
\begin{equation}
\begin{aligned}
\frac{\eta-\gamma}{\eta}= \log\left(\int \pi(a|s,\a_t)\exp(\frac{Q(s,a)}{\eta})da\right).
\label{eq:norm}
\end{aligned}
\end{equation}

Now to obtain the dual function $g(\eta)$, we plug in the solution to the KL constraint term of the lagrangian and it results in, 

\begin{equation*}
  \begin{aligned}
L(q,\eta,\gamma) =& \int \mu(s)\int q(a|s) Q(s,a) dads \\
&- \eta\int \mu(s)\int q(a|s)\Big[\frac{Q(s,a)}{\eta} + \log \pi(a|s;\a_t) - \frac{\eta-\gamma}{\eta}\Big]dads + \eta\epsilon
\\ &+\eta\int \mu(s)\int q(a|s)\log \pi(a|s;\a_t)dads + \gamma\left(1 - \iint \mu(s) q(a|s)dads\right).
  \end{aligned}
\end{equation*}

Most of the terms cancel out and after rearranging the terms we obtain,

\begin{equation*}
  \begin{aligned}
L(q,\eta,\gamma) = \eta\epsilon + \eta\int \mu(s)\frac{\eta-\gamma}{\eta}ds.
  \end{aligned}
\end{equation*}

Note that we have already calculated the term inside the integral in equation \ref{eq:norm}. By plugging in equation \ref{eq:norm} we will have the dual function,

$$g(\eta) = \eta\epsilon+\eta\int\mu(s)\log\left(\int \pi(a|s, \a_t)\exp(\frac{Q(s,a)}{\eta})da\right).$$




\section{Relation to Evolutionary Strategy algorithms}

On a high level, the difference between our algorithm and evolutionary strategy (ES) algorithms is that the problem handled by our algorithm is stateful, whereas in ES one typically considers a bandit problem. Another difference is that in ES, the value function (or in the stateless case, the reward function) does not change and the goal is to find the optimum solution given a fixed reward function. However in our setting the Q-function changes when the policy changes. Nonetheless, if we consider only a one-step policy improvement for one single and fixed state, given a Q-function -- while staying close to the old policy -- then we can recover the mean and covariance update rules of CMAES (assuming that the Gaussian policy is not parameterized by a non-linear function as in our main paper).

Concretely, considering a bandit problem, we can perform state-free optimization by directly sampling a set of actions from the old policy and evaluate them given the reward function. After that we can use any weighting method such as ranking or exponential transformation to re-weight the actions.
Subsequently we can solve the decoupled objectives in the main text section 4  in closed form when we use a soft constraint on KL in Step 3 (and, as mentioned above assuming mean $\mu$ and covariance $\Sigma$ are our only parameters), i.e, $$\mu_{new} = (1-\alpha_{\mu}) \mu_{old} + \alpha_{\mu} \sum_i w_i a_i$$
$$\Sigma_{new} = (1-\alpha_{\Sigma}) \Sigma_{old} + \alpha_{\Sigma} \sum_i w_i (a_i-\mu_{old})(a_i-\mu_{old})^T.$$
Here $\alpha_{\mu}$ and $\alpha_{\Sigma}$ define how much we move from the old distribution. This is the exact update rule for CMA-ES with the difference that CMA-ES sets $\alpha_{\mu}$ to zero resulting in an unregulated update rule for the mean of the Gaussian policy. This choice makes sense when one is optimizing against the true reward function. However, in the reinforcement learning setting the Q-function should be estimated and typically has high variance. Therefore a constraint on the mean to limit exploiting the Q-function is necessary as we showed in our experiments. Please see \citep{abdolmaleki2017TRCMA,abdolmaleki2017contextual} for more details on derivations.

The above is not only the case for CMA-ES. Depending on the weighing strategy, the interpolation factors $\alpha_{\mu}$,$\alpha_{\Sigma}$ and the use of $\mu_{old}$ or $\mu_{new}$, we recover the update rules not only for CMA-ES, but also Episodic PI$^2$, Episodic PI$^2$-CMA, Episodic Reps, Episodic Power, Cross Entropy methods and EDAs \cite{Deisenroth2013}. We recover TR-CMAES \cite{abdolmaleki2017TRCMA} in the case where instead of the soft constraints in M-step, we use the hard constraints on the KL. If we use our formulation for contextual RL with a linear function approximator and state-independent covariance we recover the update rules from contextual CMA-ES \cite{abdolmaleki2017contextual}.  

One interesting observation is that the per state solution we obtain (assuming no generalization over states is performed via a neural network), is a convex interpolation between the last policy and the sample based  policy. The change of this distribution is upper bounded for each state, i.e, in case of a Gaussian distribution the policy for each state is at most the sample Gaussian distribution, even when we set the constraint on the KL to infinity. As a matter of fact, the current policy is changing towards the sample policy in Step 3. This can be interpreted as following a natural gradient where the maximum change is upper bounded and the direction of the change is the optimal improved distribution.

\section{Ranking versus Exponential Transformation}

\begin{figure*}[ht]
\centering
\begin{minipage}[c]{1\textwidth}
\def\mywidth{1.0}
\def\myhsep{-0.01}
\includegraphics[width=\mywidth\textwidth]{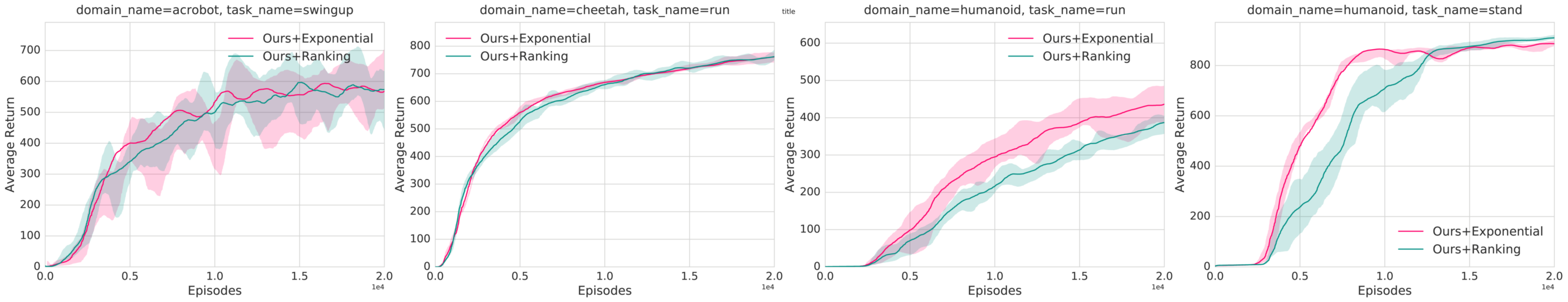}

\caption{This plot compares two different weighting methods: ranking and exponential transformation. Interestingly both methods perform similarly well and we did not observe a noticeable difference.}
\label{fig:ranking}
\end{minipage}
\end{figure*}

Figure \ref{fig:ranking} compares two different strategies for weighting actions, exponential transformation and ranking. For the ranking results we weight the actions for each state using the following formula:
$$w_{i} \propto \ln (\frac{N + \eta}{i}),$$ where $i$ is the rank of the action based on its Q-value, N is number of actions per state (which is 20 in our case) and $\eta$ is temperature parameter which we set to 10.  

We did not observe a noticeable difference between the two transformations. However, we recommend an exponential transformation over ranking. Mainly because it allows for efficient optimization of the temperature parameter. 

\section{Additional Visualizations Regarding Premature Convergence}

\begin{figure*}[ht]
\centering
\begin{minipage}[c]{1\textwidth}
\def\mywidth{0.24}
\def\myhsep{-0.01}

\begin{subfigure}[b]{\textwidth}
\includegraphics[width=\mywidth\textwidth]{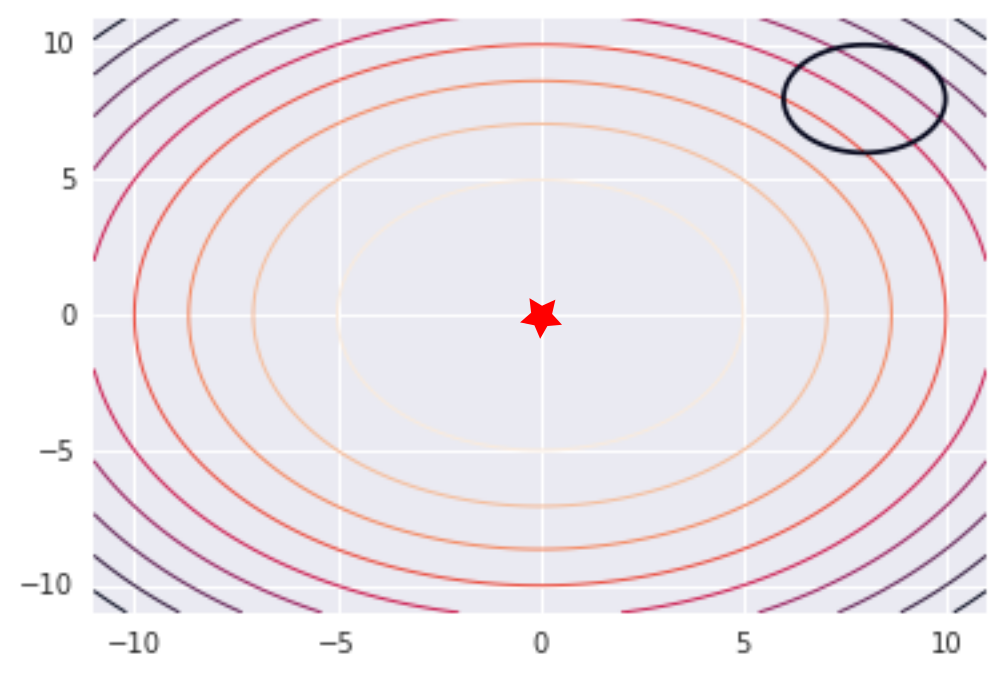}
\includegraphics[width=\mywidth\textwidth]{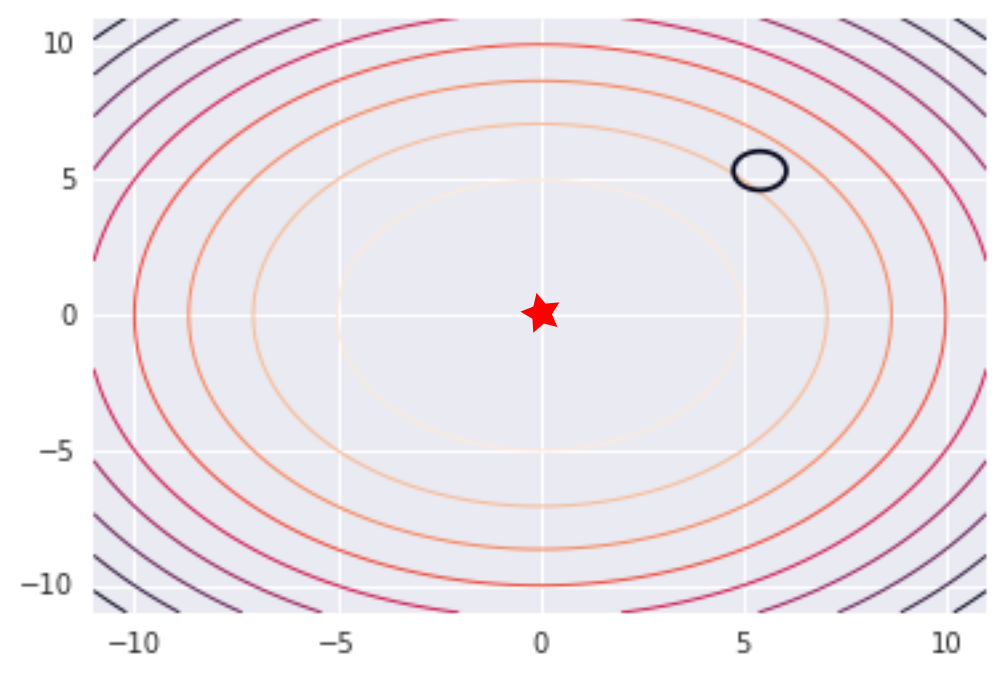}
\includegraphics[width=\mywidth\textwidth]{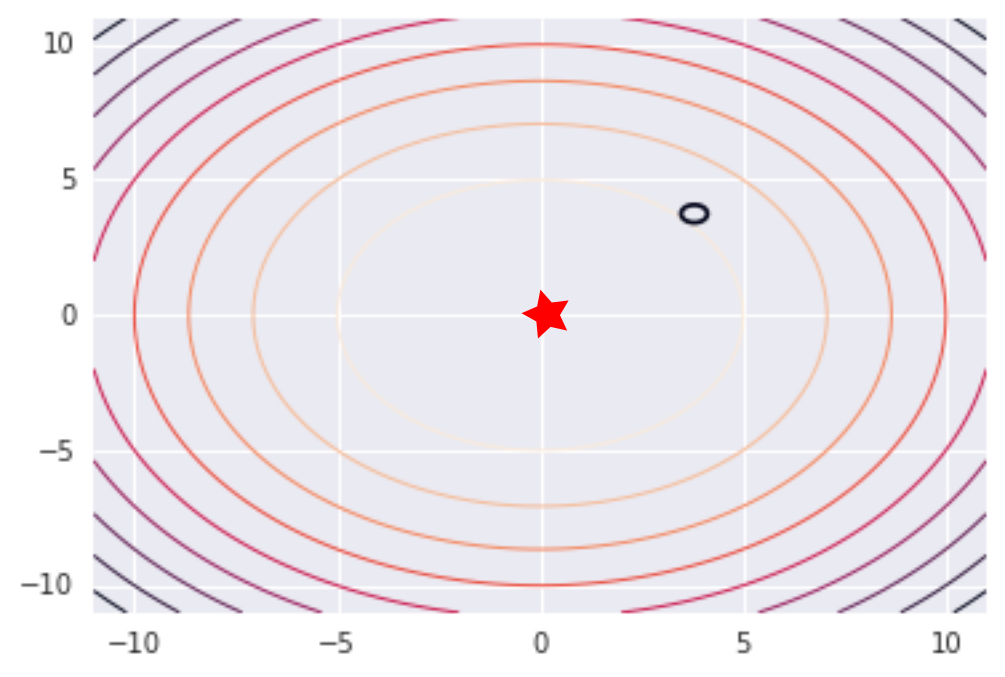}
\includegraphics[width=\mywidth\textwidth]{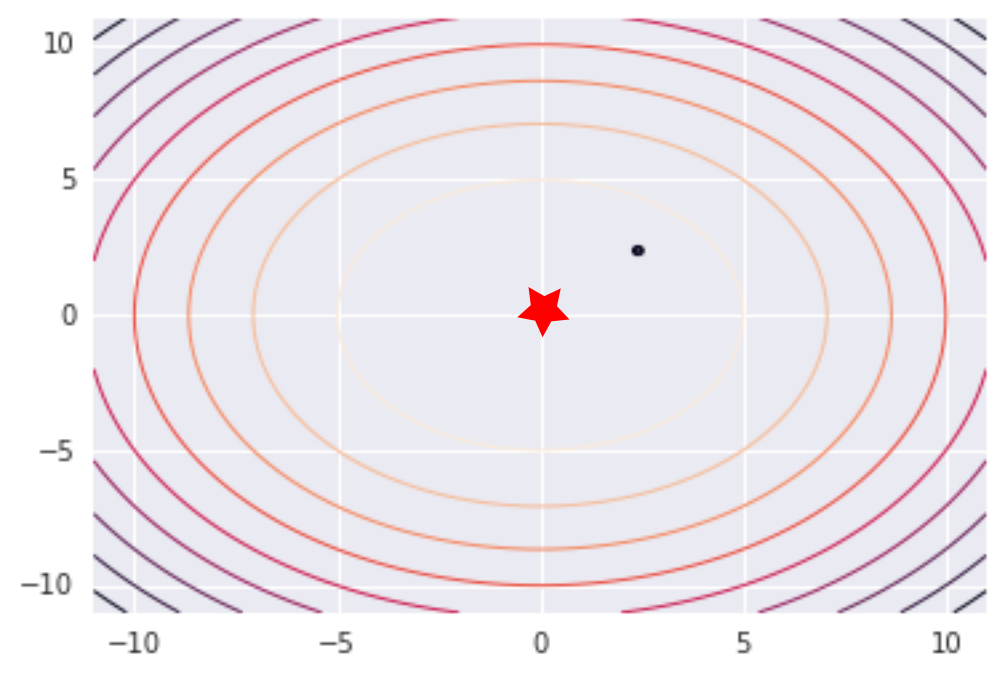}
\caption{Maximum Likelihood Loss}
\end{subfigure}

\begin{subfigure}[b]{\textwidth}
\includegraphics[width=\mywidth\textwidth]{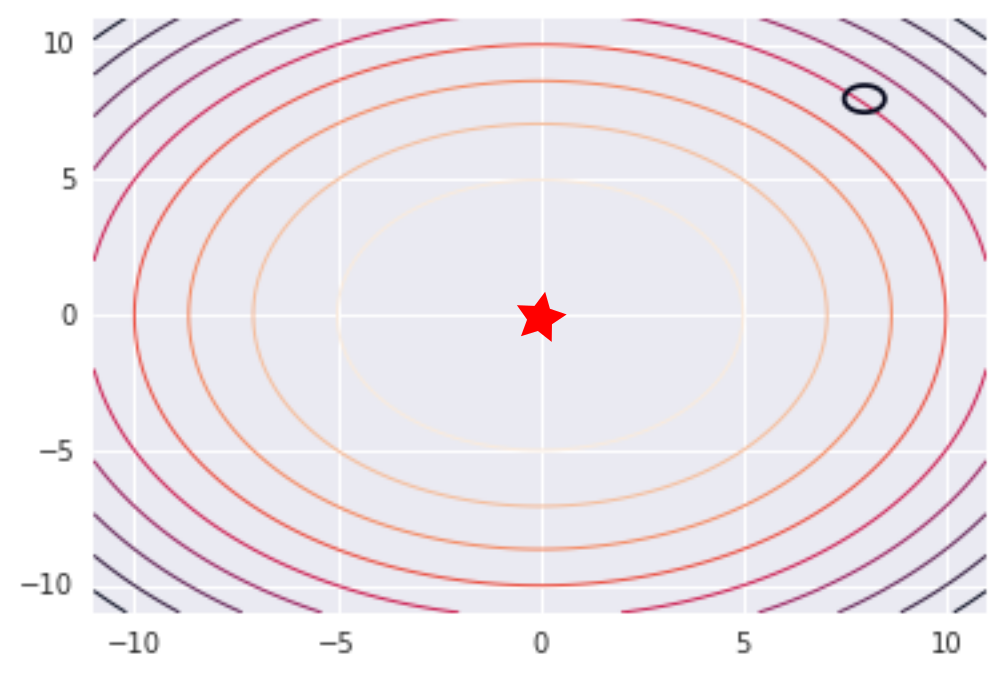}
\includegraphics[width=\mywidth\textwidth]{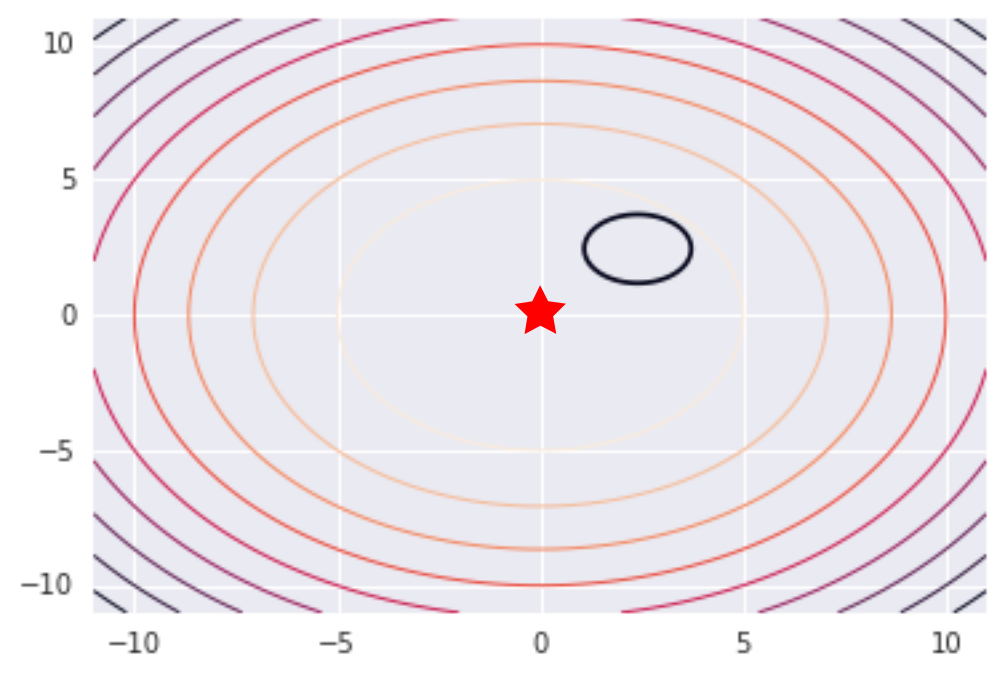}
\includegraphics[width=\mywidth\textwidth]{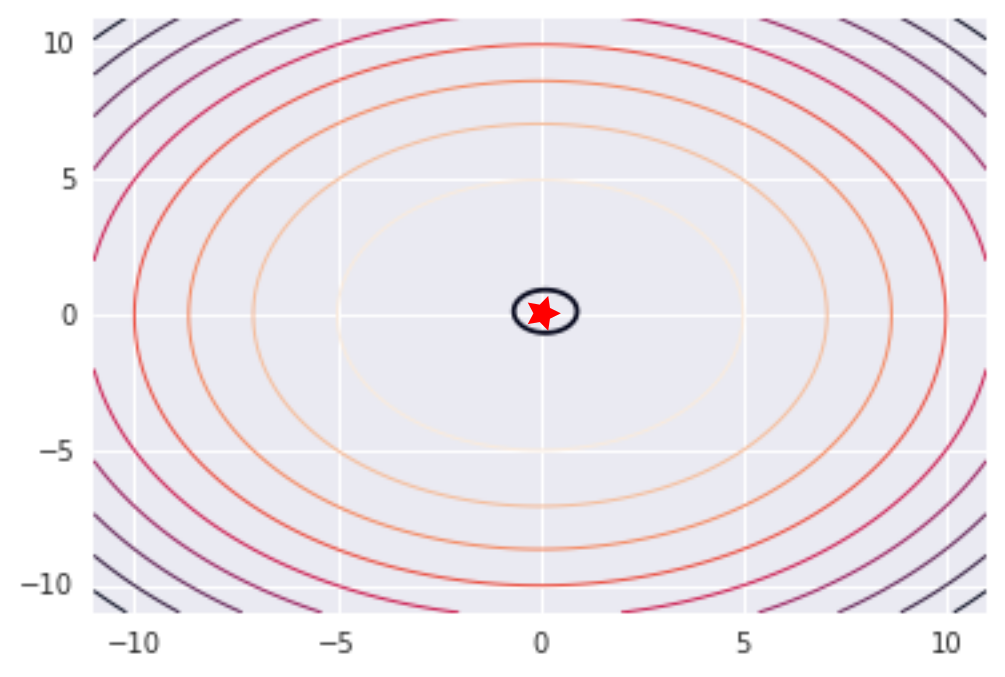}
\includegraphics[width=\mywidth\textwidth]{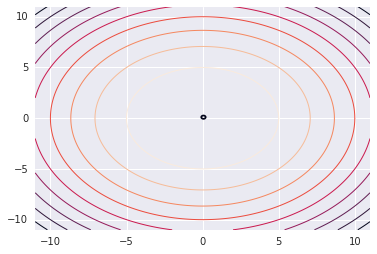}
\caption{Decoupled Maximum Likelihood Loss (See Section 4 in main text)}
\end{subfigure}
\caption{We plot the landscape of a quadratic Q-function for state [0,0]. The black circle is the Gaussian policy for state [0,0] in each iteration. From left to right we depict policy updates for state [0,0] (20 gradient steps between each plot). The first row shows the results when maximum likelihood estimation from is used, which leads to reducing variance and premature convergence. The second row shows results when the decoupled maximum likelihood objective is used. In this case, the optimum can be found reliably, even when we start with a small initial policy distribution. We can see that the variance first naturally grows and then shrinks again to pin down the optimum.}
\label{fig:landscape}
\end{minipage}
\end{figure*}

Figure \ref{fig:landscape} visualizes the evolution of the policy for state [0,0] when optimizing a statefull Q function that is quadratic in action space. The results show that when we use MLE, without decoupling the updates for mean and covariance, the policy suffers from premature convergence. However, when we decouple the updates the variance naturally grows and shrinks.   

\section{Additional Experiments on DeepMind control suite tasks}
Figure \ref{fig:additionalExperiments} provides additional results on the control suite tasks, comparing against two baselines.

\section{Experiment details}

In this section we outline the details on the hyper-parameters used for our algorithm and baselines, DDPG and SVG. All continuous control experiments use a feed-forward neural network except for Parkour tasks as described in section \ref{sec:parkour-network}. The policy is given by a Gaussian distribution with a diagonal covariance matrix, i.e, 
$
  \pi(\vec a| \vec s,\a) = \mathcal{N}\left(\mu , \vec
  \Sigma \right)
$. The neural network outputs the mean $\mu=\mu(s)$ and diagonal Cholesky factors $A=A(s)$, such that $\Sigma = AA^T$. The diagonal factor $A$ has positive diagonal elements enforced by the softplus transform $A_{ii} \leftarrow \log(1 + \exp(A_{ii}))$ to enforce positive definiteness of the diagonal covariance matrix.

Tables \ref{t:MPO},\ref{t:SVG} and \ref{t:DDPG} show the hyper parameters we used for all three algorithms. We found layer normalization and tanh on output of the layer normalization are important for stability of all algorithms. 

We also found that: 1) a tanh operation on the mean of the distribution and 2) forcing a minimum variance are required for DDPG and SVG. We emphasize that our algorithm does not use any such tricks. 

For our algorithm the most important hyper parameters are the constraints in Step 1 and Step 2.


\begin{table}[t]
\begin{center}
 \begin{tabular}{c||c} 
 Hyperparameters & SVG \\
 \hline
 Policy net & 200-200-200\\ 
 Q function net & 500-500-500\\
 Entropy Regularization Factor & 0.001\\
 Discount factor ($\gamma$) & 0.99 \\
 Adam learning rate & 0.0003 \\
 Replay buffer size & 2000000 \\
 Target network update period & 250 \\
 Batch size & 3072\\
 Activation function & elu\\
 Tanh on output of layer norm & Yes\\
 Layer norm on first layer & Yes\\
 Tanh on Gaussian mean & Yes \\
 Min variance & 0.1\\
 Max variance & unbounded 
\end{tabular}
\end{center}
\caption{Hyper parameters for SVG}
\label{t:SVG}
\end{table}

\subsection{Settings for standard functions}
We use a two layer neural network with 50 neurons to map the current state of the network to the mean and diagonal covariance of the Gaussian policy. The parameters of this neural network $\a$ are then optimized using the procedure described in the algorithm section. 
We set $\epsilon=0.1$, $\epsilon_{\mu} = 5.0$ and
$\epsilon_{\Sigma} = 0.001$. Please note that, the mean is effectively unregulated because of a loose KL bound. The reason is that, here we have access to a perfect Q-function and therefore we can exploit it as much as it is possible. This situation is different in the RL setting, where the Q-function is estimated and can be noisy.

\begin{table}[t]
\begin{center}
 \begin{tabular}{c||c} 
 Hyperparameters & Ours \\
 \hline
 Policy net & 200-200-200\\ 
 Number of actions sampled per state& 20\\
 Q function net & 500-500-500\\
 $\epsilon$ & 0.1 \\
 $\epsilon_{\mu}$ & 0.0005 \\
 $\epsilon_{\Sigma}$ & 0.00001\\
 Discount factor ($\gamma$) & 0.99 \\
 Adam learning rate & 0.0003 \\
 Replay buffer size & 2000000 \\
 Target network update period & 250\\
 Batch size & 3072\\
 Activation function & elu\\
 Layer norm on first layer & Yes\\
 Tanh on output of layer norm & Yes\\
 Tanh on Gaussian mean & No \\
 Min variance & Zero\\
 Max variance & unbounded 
\end{tabular}
\end{center}
\caption{Hyper parameters for our algorithm with decoupled update on mean and covariance}
\label{t:MPO}
\end{table}

\subsection{Additional Details on the SVG baseline}
For the stochastic value gradients (SVG-0) baseline we use the same policy parameterization as for our algorithm, e.g. we have 
$$
\pi_\theta = \mathcal{N}(\mu_\theta(s),\sigma^2_\theta(s) I),
$$
where $I$ denotes the identity matrix and $\sigma_\theta(s)$ is computed from the network output via a softplus activation function. 

To obtain a baseline that is, in spirit, similar to our algorithm we used SVG in combination with Entropy regularization. That is, we optimize the policy via gradiend ascent, following the reparameterized gradient for a given state s sampled from the replay:
\begin{equation}
\begin{aligned}
\nabla_\theta \mathbb{E}_{\pi_\theta(a | s)}[Q(a, s)] + \alpha \mathrm{H}\Big(\pi_\theta(a | s)\Big),
\end{aligned}
\end{equation}
which can be computed, using the reparameterization trick, as 
\begin{equation}
\begin{aligned}
 \mathbb{E}_{\zeta \sim \mathcal{N}(0, I)}[\nabla_\theta g_\theta(s, \zeta) \nabla_g Q(g_\theta(s, \zeta), s)] + \alpha \nabla_\theta \mathrm{H}\Big(\pi_\theta(a | s)\Big),
\end{aligned}
\end{equation}
where $g_\theta(s, \zeta) = \mu_\theta(s) + \sigma_\theta(\bs) * \zeta$ is now a deterministic function of a sample from the standard multivariate normal distribution. See e.g. \cite{svg} (for SVG) as well as \cite{rezende14,kingma2013auto} (for the reparameterization trick) for a detailed explanation.

\subsection{Additional Details on the DDPG baseline}
For the DDPG baseline we parameterize only the mean of the policy $\mu_\theta(x)$, using fixed univariate Gaussian exploration noise of $\sigma = 0.3$ in all dimensions, that is the behaviour policy for collecting experience can be described as 
$$
\pi_{\text{DDPG}}(a | s) = \mathcal{N}(\mu_\theta(x), 0.3 * I),
$$
where $I$ denotes the identity matrix. To improve the mean of the policy we follow the deterministic policy gradient:
\begin{equation} \mathbb{E}_{\zeta \sim \mathbb{N}(0, 0.3 * I)}\Big[ \nabla_\theta \mu_\theta(s) \nabla_a Q(a, s) \mid a = \mu_\theta(s) + \zeta  \Big].
\end{equation}
Policy evaluation is performed with Q-learning as described in the main text. The hyperparameters used for DDPG are described in Table \ref{t:DDPG}. We highlight that good performance for DDPG can be achieved when hyperparameters are tuned correctly; we found that a tanh activation function on the mean combined with layer normalization in the first layer of policy and Q-function are crucial in this regard.

\begin{table}[ht]
\begin{center}
 \begin{tabular}{c||c} 
 Hyperparameters & DDPG \\
 \hline
 Policy net & 200-200-200\\ 
 Q function net & 500-500-500\\
 Discount factor ($\gamma$) & 0.99 \\
 Adam learning rate & 0.0001 \\
 Replay buffer size & 2000000 \\
 Target network update period & 250\\
 Batch size & 3072\\
 Activation function & elu\\
 Tanh on networks input & Yes\\
 Tanh on output of layer norm & Yes\\
 Tanh on Gaussian mean & Yes \\
 Min variance & 0.3\\
 Max variance & 0.3 
\end{tabular}
\end{center}
\caption{Hyper parameters for DDPG}
\label{t:DDPG}
\end{table}
\subsection{Network Architecture for Parkour}
\label{sec:parkour-network}
For the parkour experiments we used the same hyperparameters but changed the architecture of the feed-forward network. This is due to the fact that for these problems both proprioceptive information about the robot's state as well as information of the terrain height is available. We thus used the same network architecture as in \cite{d4pg,abdolmaleki2018maximum}, which in turn was derived from the networks in \cite{heess2017emergence}: each of the two input streams is passed through a two-layer feed-forward neural network with 200 units each for the critic (100 units each for the actor), before being passed through one layer of 100 units combining both modalities; on top of which a final layer computes the Q-value (or in case of the policy produces mean and diagonal covariance). 

\section{Additional Experiments on the Control Suite}

\begin{figure*}[h]
\centering
\begin{minipage}[c]{1\textwidth}
\def\mywidth{0.18}
\includegraphics[width=\mywidth\textwidth]{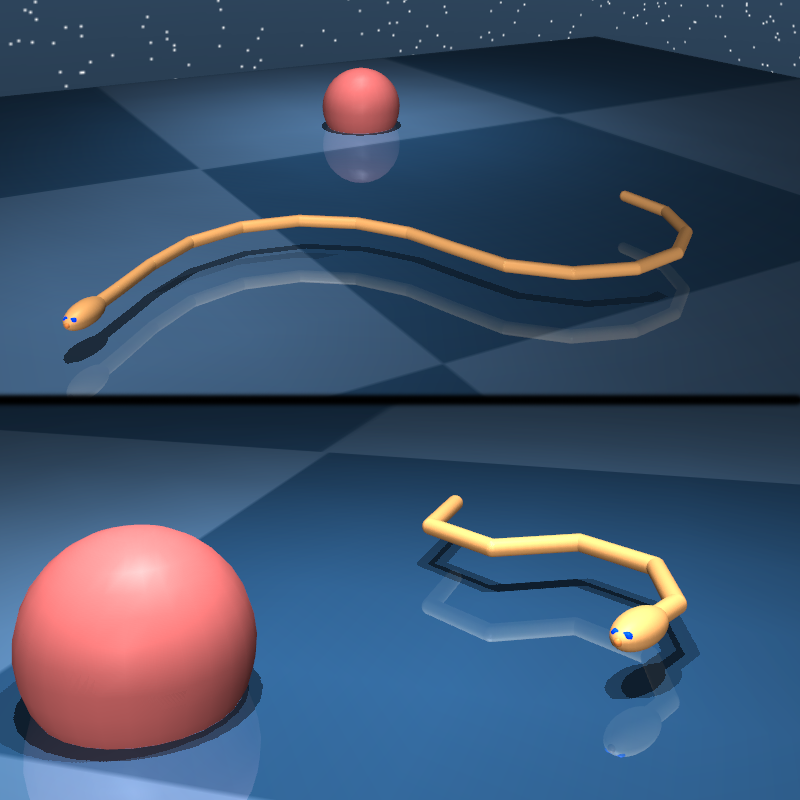}
\includegraphics[width=\mywidth\textwidth]{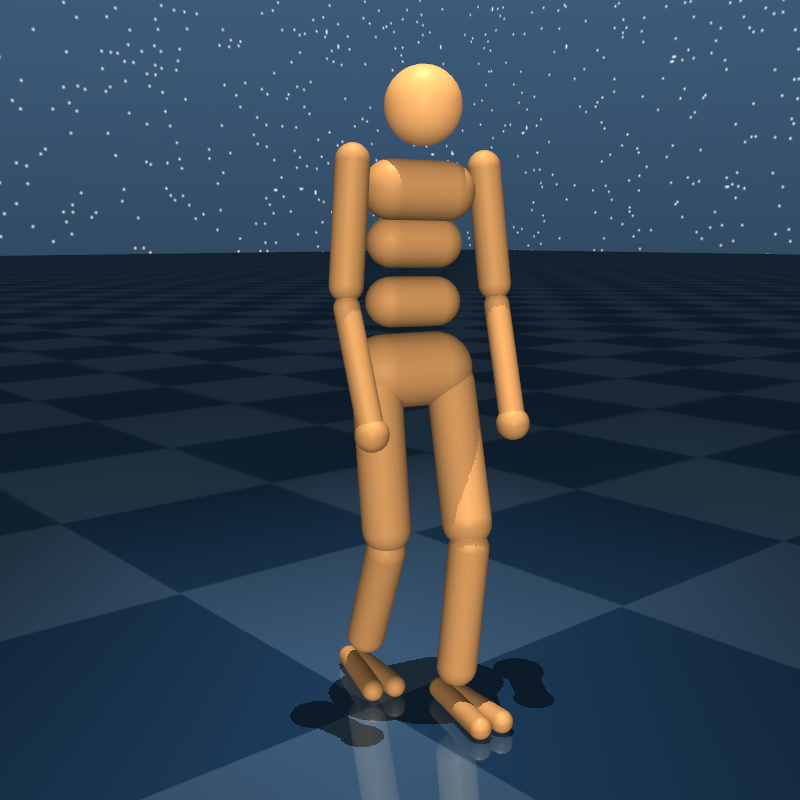}
\includegraphics[width=\mywidth\textwidth]{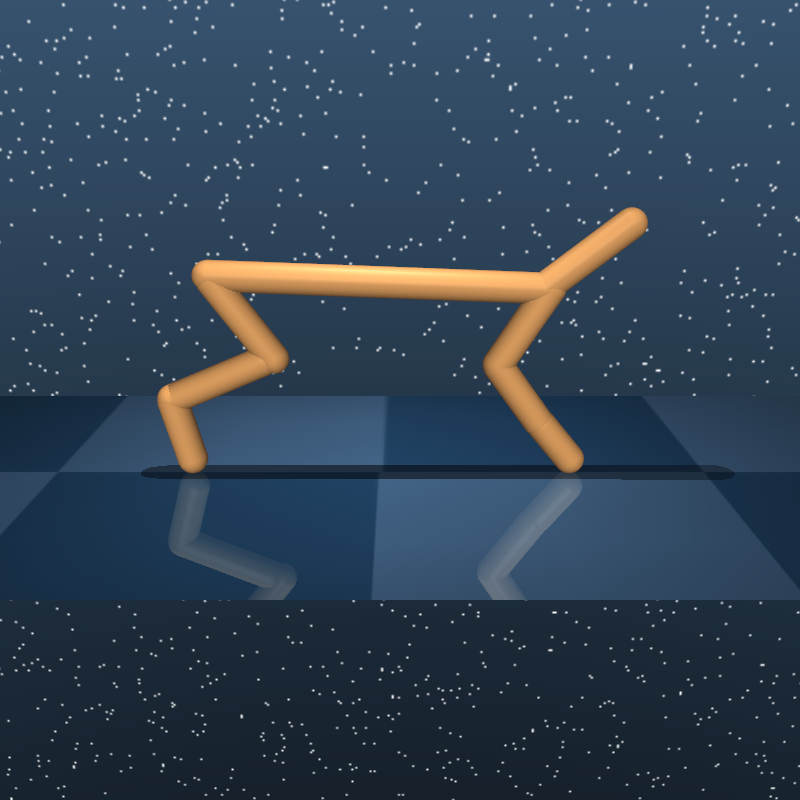}
\includegraphics[width=\mywidth\textwidth]{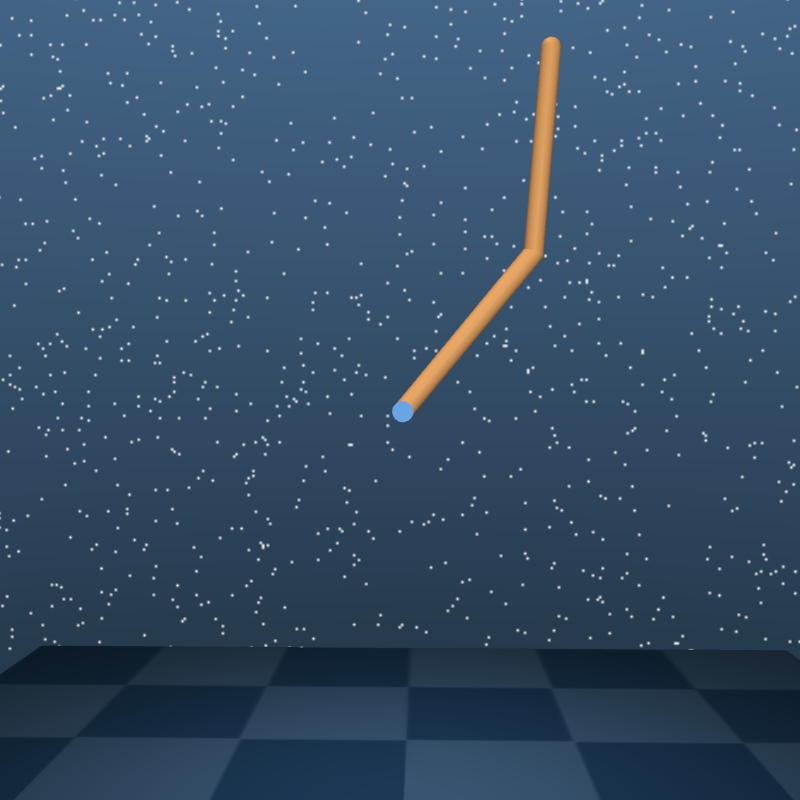}
\includegraphics[width=\mywidth\textwidth]{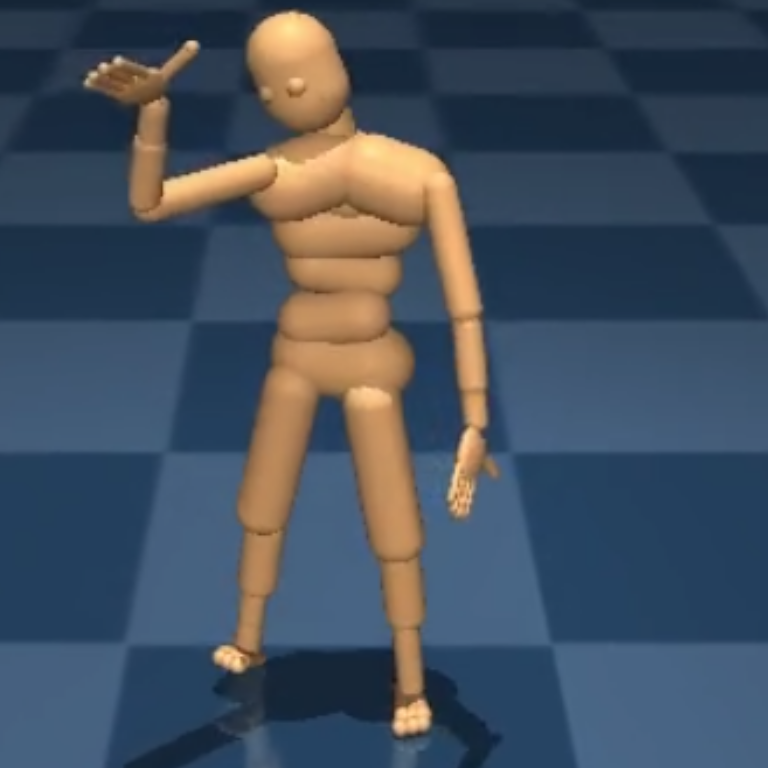}
\caption{From left to right, depiction of swimmer, humanoid , cheetah, acrobat and humanoid-cmu domains from the DeepMind control suite.}
\label{fig:controlsuite}
\end{minipage}
\end{figure*}

\begin{figure*}[h]
\centering
\begin{minipage}[c]{1\textwidth}
\def\mywidth{1.0}
\includegraphics[width=\mywidth\textwidth]{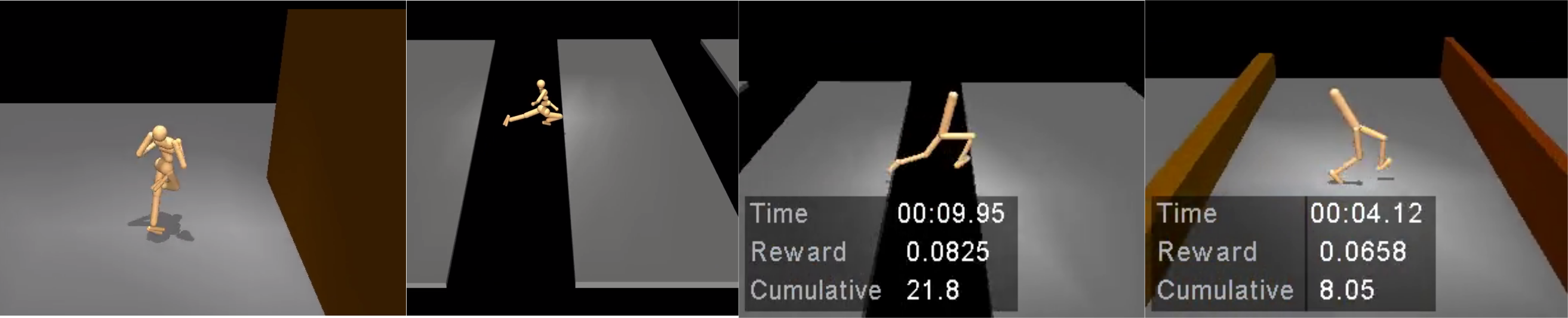}
\caption{We use three high dimensional control problems from \citet{heess2017emergence} for evaluation: 1- Parkour 3D-walls and Parkour 3D-gaps with 22 dimensional action space and 539 dimensional state space (Left images) 2-Parkour-2D with 6 dimensional action space and 120 dimensional state space (Right images)}
\label{fig:parkour}
\end{minipage}
\end{figure*}

We provide a full evaluation on 27 tasks from the DeepMind control suite (see Figure \ref{fig:controlsuite}) and the parkour suite (see Figure \ref{fig:parkour}). Please see the main text for the results on parkour tasks. Figure \ref{fig:svg-mpo} and \ref{fig:additionalExperiments} in appendix shows the full results for the control suite tasks. The results suggest that while other baselines perform well, only our algorithm performs well across all tasks, achieving better asymptotic performances for high dimensional tasks.

\begin{figure*}[ht]
\centering
\begin{minipage}[c]{1\textwidth}
\def\mywidth{0.33}
\def\myhsep{-0.01}
\includegraphics[width=\mywidth\textwidth]{./figures/svg-plots/acrobot_swingup_svg_mpo2.pdf}
\hspace{\myhsep\textwidth}
\includegraphics[width=\mywidth\textwidth]{./figures/svg-plots/swimmer15_svg_mpo.pdf}
\hspace{\myhsep\textwidth}
\includegraphics[width=\mywidth\textwidth]{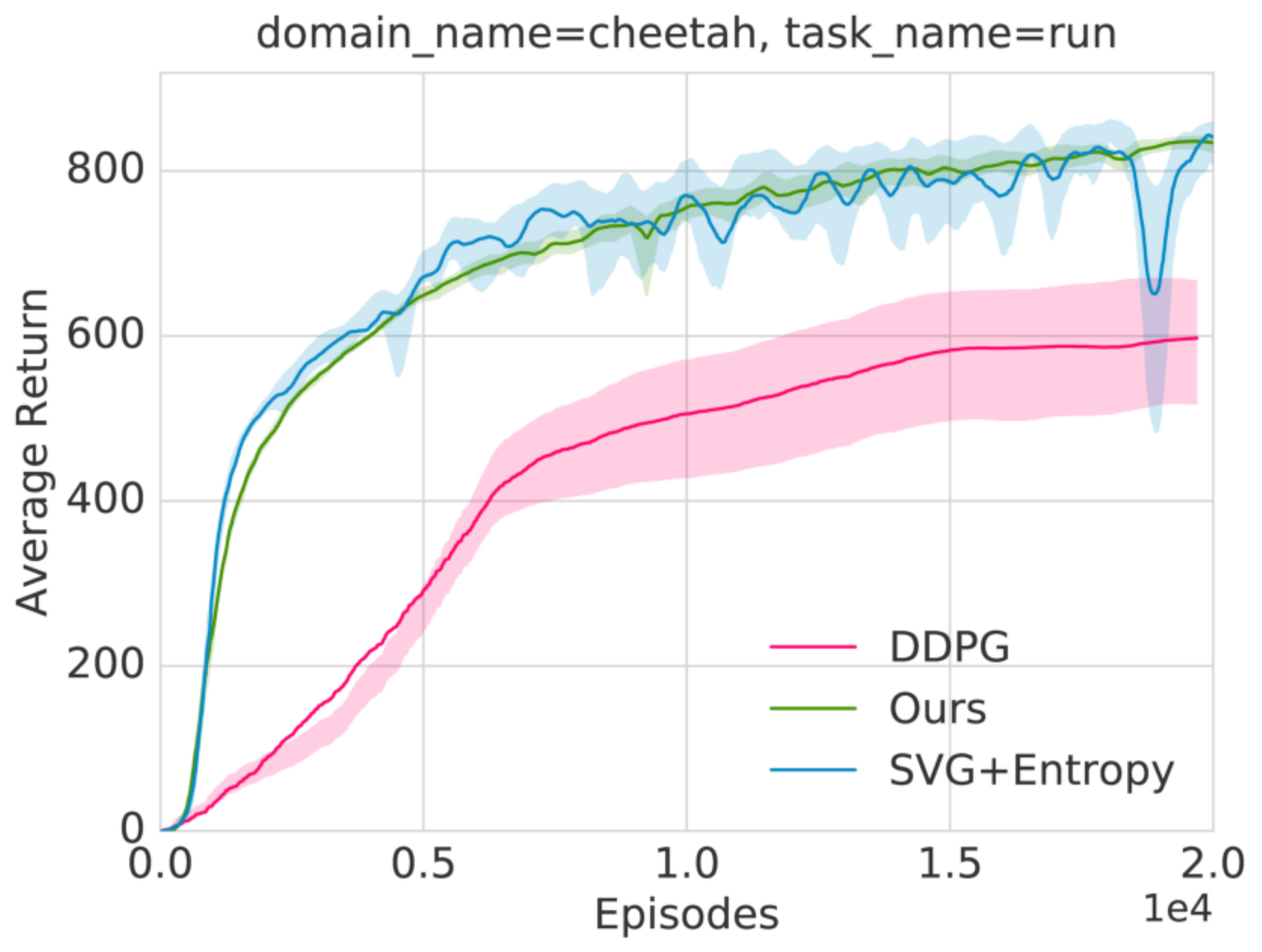}
\hspace{\myhsep\textwidth}\\
\vspace{-0.027\textwidth}

\hspace{\myhsep\textwidth}
\includegraphics[width=\mywidth\textwidth]{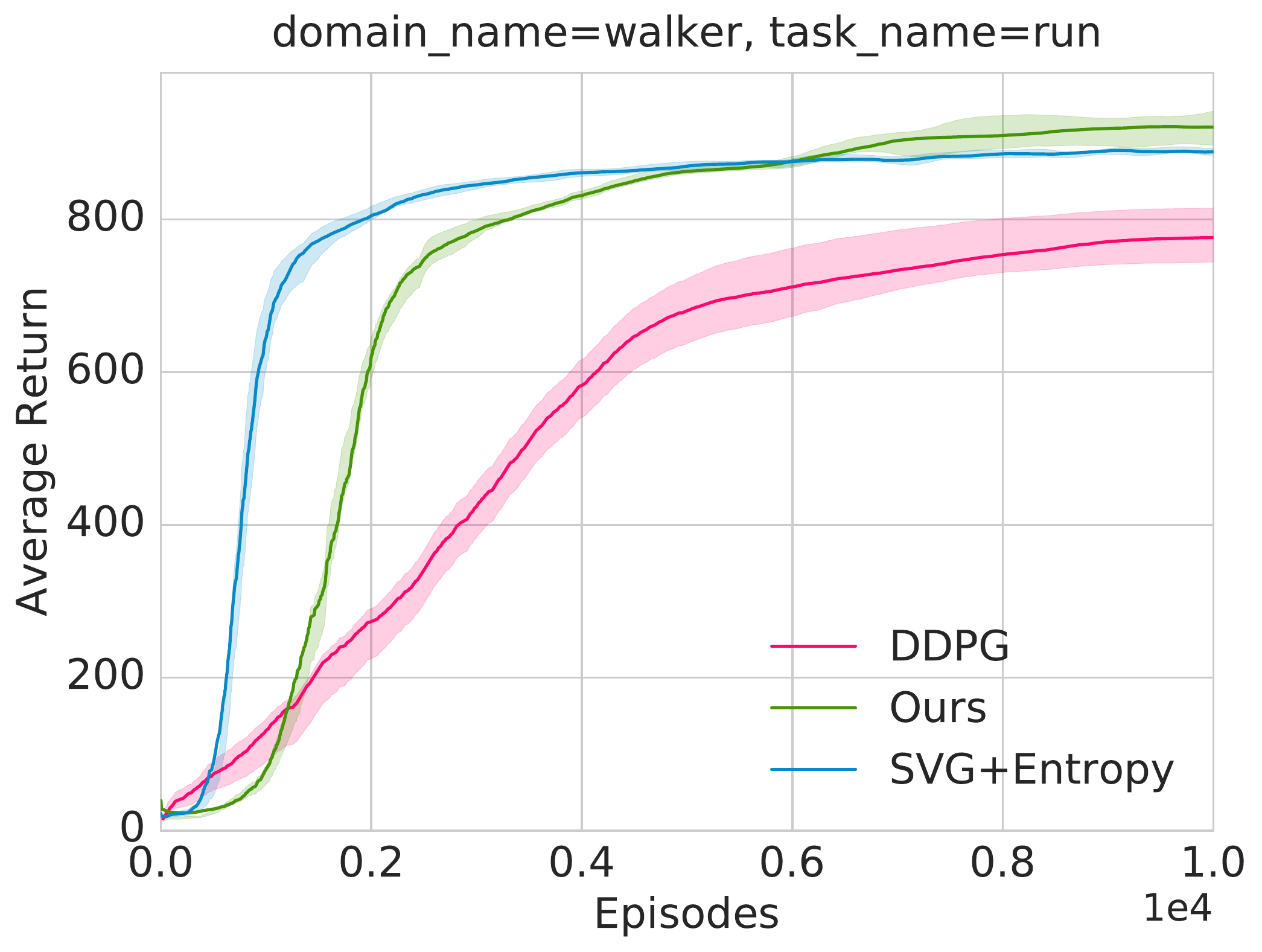}
\hspace{\myhsep\textwidth}
\includegraphics[width=\mywidth\textwidth]{./figures/svg-plots/humanoid_run_svg_mpo.pdf}
\hspace{\myhsep\textwidth}
\includegraphics[width=\mywidth\textwidth]{./figures/svg-plots/cmustand_svg_mpo.pdf}

\caption{Comparison of our algorithm against SVG and DDPG. Results show that while all algorithms perform similar in low-dimensional tasks (Acrobot, Swimmer, cheetah and walker) when hyperparameters are tuned correctly; differences start to emerge in high dimensional tasks i.e, humanoid run (22 actions) and humanoid CMU-stand(56 action dimension) where our algorithm performs more stable and with better asymptotic performance.}
\label{fig:svg-mpo}
\end{minipage}
\end{figure*}

\begin{figure*}[ht]
\centering
\begin{minipage}[c]{1\textwidth}
\def\mywidth{0.9}
\def\myhsep{-0.01}
\includegraphics[width=\mywidth\textwidth]{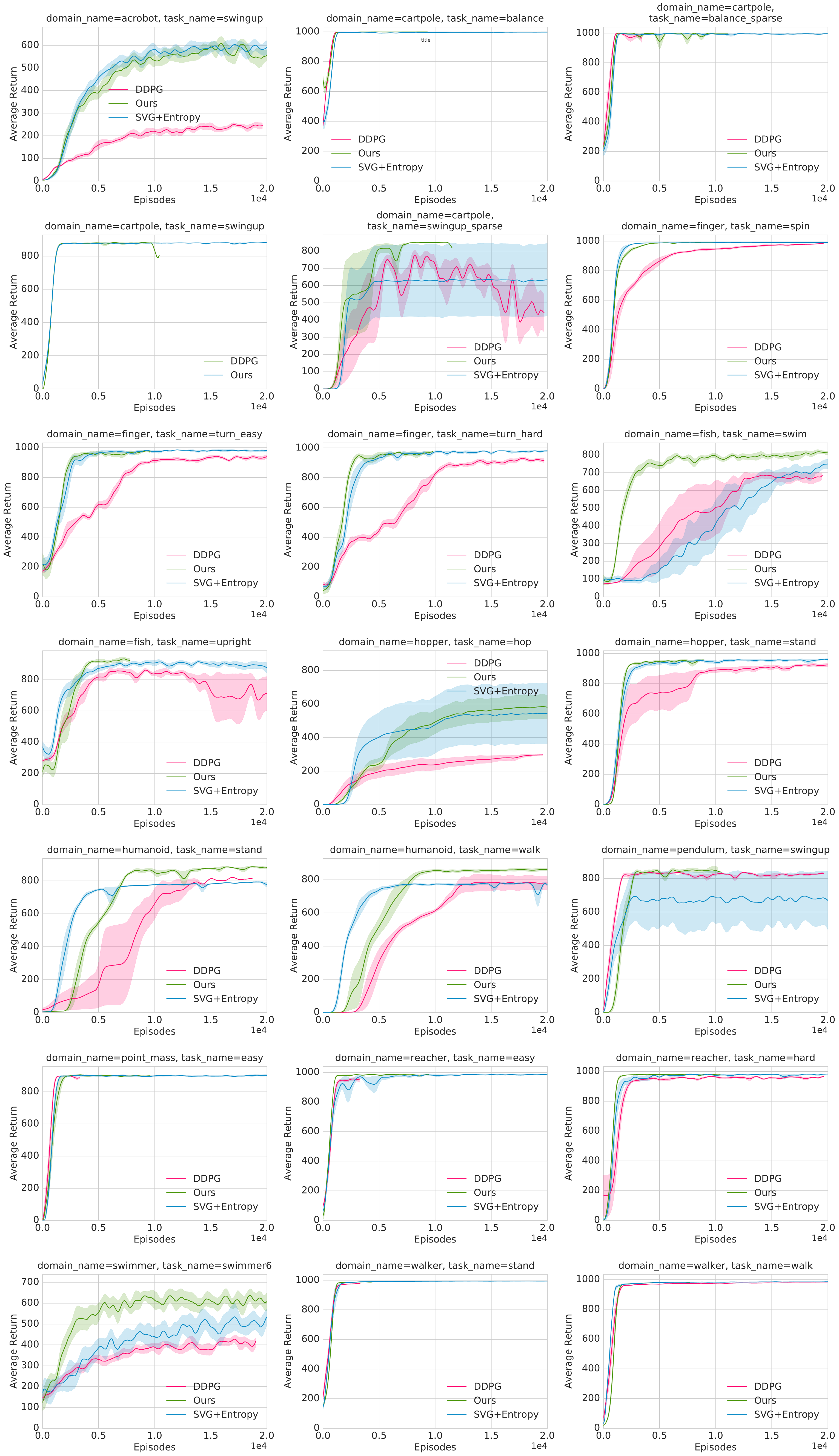}

\caption{In this plot we show the results on solving additional DeepMind control suit tasks}
\label{fig:additionalExperiments}
\end{minipage}
\end{figure*}

\end{document}